\documentclass{article}

 \PassOptionsToPackage{numbers, compress}{natbib}

\usepackage[preprint]{neurips_2025}

\usepackage[utf8]{inputenc} 
\usepackage[T1]{fontenc}    
\usepackage{hyperref}       
\usepackage{url}            
\usepackage{booktabs}       
\usepackage{amsfonts}       
\usepackage{nicefrac}       
\usepackage{microtype}      
\usepackage{xcolor}         

\usepackage{subcaption}
\usepackage{array}
\usepackage{algorithm}
\usepackage{algorithmic}
\usepackage{amsthm}
\usepackage{amsmath}
\usepackage{amssymb}
\usepackage{mathtools}
\usepackage{wrapfig}
\usepackage{multirow}
\usepackage{epstopdf}
\usepackage{makecell}
\usepackage{graphicx}
\usepackage{appendix}
\usepackage{bm}
\usepackage{bbm}
\usepackage{soul,color}
\usepackage{adjustbox}
\usepackage[normalem]{ulem}
\usepackage{colortbl}
\usepackage{tikz}
\usepackage{tikz-cd}
\usetikzlibrary{arrows.meta, decorations.pathreplacing, fit, shapes.geometric}
\usepackage[capitalize,noabbrev]{cleveref}

\theoremstyle{plain}
\newtheorem{theorem}{Theorem}[section]
\newtheorem{proposition}[theorem]{Proposition}

\theoremstyle{definition}
\newtheorem{definition}[theorem]{Definition}

\theoremstyle{remark}

\usepackage[textsize=tiny]{todonotes}

\newcommand{\method}{\textsc{SCR}\xspace} 

\newcommand{\graph}{\mathcal{G}}
\newcommand{\gtrain}{\mathcal{G}_{\textit{tr}}}
\newcommand{\ginf}{\mathcal{G}_{\textit{inf}}}

\newcommand{\etrain}{\mathcal{E}_{\textit{tr}}}
\newcommand{\einf}{\mathcal{E}_{\textit{inf}}}
\newcommand{\rtrain}{\mathcal{R}_{\textit{tr}}}
\newcommand{\rinf}{\mathcal{R}_{\textit{inf}}}
\newcommand{\ttrain}{\mathcal{T}_{\textit{tr}}}
\newcommand{\tinf}{\mathcal{T}_{\textit{inf}}}

\newcommand{\ents}{\mathcal{E}}
\newcommand{\rels}{\mathcal{R}}
\newcommand{\features}{\mathcal{X}}
\newcommand{\triples}{\mathcal{T}}

\usepackage{amsmath,amsfonts,bm}
\usepackage{wrapfig}
\usepackage{tabularx}
\usepackage{xspace}
\usepackage[shortlabels]{enumitem}
\theoremstyle{plain}

\newcommand{\updatefunc}{\small \textsc{Upd}\xspace}
\newcommand{\aggregate}{\small\textsc{Agg}\xspace}

\newcommand{\mes}{\small\textsc{Msg}\xspace}
\newcommand{\init}{\small\textsc{Init}\xspace}

\def\eqref#1{equation~\ref{#1}}

\def\1{\bm{1}}

\newcommand{\rdwl}{{\sf rawl}}

\newcommand{\rwl}{{\sf rwl}}

\def\vh{{\bm{h}}}

\def\vr{{\bm{r}}}

\DeclareMathAlphabet{\mathsfit}{\encodingdefault}{\sfdefault}{m}{sl}
\SetMathAlphabet{\mathsfit}{bold}{\encodingdefault}{\sfdefault}{bx}{n}

\title{Towards Graph Foundation Models: \\ Training on Knowledge Graphs Enables Transferability to General Graphs}

\author{
Kai Wang$^{\dag}$, Siqiang Luo$^\dag$\thanks{~~The Corresponding Author}, \ Caihua Shan$^{\ddag}$, Yifei Shen$^{\ddag}$\\
 $^{\dag}$Nanyang Technological University \ 
 $^{\ddag}$Microsoft Research Asia \\
\texttt{kai\_wang@ntu.edu.sg, siqiang.luo@ntu.edu.sg}\\
\texttt{caihuashan@microsoft.com, yifeishen@microsoft.com}\\
}

\begin{document}

\maketitle

\begin{abstract}
Inspired by the success of large language models, there is a trend toward developing graph foundation models to conduct diverse downstream tasks in various domains. 
However, current models often require extra fine-tuning to apply their learned structural and semantic representations to new graphs, which limits their versatility.
Recent breakthroughs in zero-shot inductive reasoning on knowledge graphs (KGs), offer us a new perspective on extending KG reasoning to general graph applications.
In this paper, we introduce \method, a unified graph reasoning framework designed to train on knowledge graphs and effectively generalize across a wide range of graph tasks and domains.
We begin by designing the task-specific KG structures to establish a unified topology for different task formats. 
Then we propose semantic-conditioned message passing, a novel mechanism addressing the inherent semantic isolation in traditional KG reasoning, by jointly modeling structural and semantic invariance patterns in graph representations.
To demonstrate the effectiveness, we evaluate the inductive reasoning capability of \method using 38 diverse graph datasets, covering node-level, link-level, and graph-level tasks across multiple domains. 
Our results show substantial performance gains over existing foundation models and supervised baselines, highlighting the efficacy and adaptability of our approach.
\end{abstract}

\section{Introduction}

In pursuit of artificial general intelligence, graph foundation models (GFMs) are designed to pretrain on large-scale graph data, learn generalizable representations, and adapt them to a wide range of downstream tasks~\citep{GFMSurvey-arxiv24, GFMSurvey-arxiv23}. 
However, most GFMs still face challenges, including format mismatches between pretraining objectives and downstream tasks, and semantic discrepancies between source and target datasets. As a result, extensive fine-tuning is often required.

In contrast to homogeneous or heterophilic graphs, which define a single relation among nodes, knowledge graphs capture complex, multi-relational connections among entities.
The most common task is KG reasoning, also known as KG completion, which involves learning embeddings of entities and relations to infer missing components in triples (head entity, relation, tail entity)~\citep{TransE,RotatE}. 
Beyond traditional transductive KG reasoning, recent studies~\citep{ultra,KG-ICL} utilized NBFNet~\citep{NBF-NIPS21} to enable zero-shot inductive KG reasoning. This approach learns relative representations of relations and entities conditioned on the graph structure—without any fine-tuning—enabling generalization to unseen KGs and inference of entities and relations never encountered during training~\citep{OurProLINK}.

Inspired by these breakthroughs, we improve the transferability of GFMs from a novel perspective: pre-training on knowledge graphs using inductive reasoning as the training objective, and then transferring to other graph domains, such as citation and molecular graphs, to perform downstream tasks like node and graph classification. Nevertheless, developing such a GFM faces two significant problems. First, it is challenging to generalize the KG reasoning format across diverse graph tasks and transfer learned representations to general graphs effectively. Second, while KG inductive reasoning excels at capturing structural patterns, it often fails to incorporate node features. Integrating node features into the learning process and enhancing pattern learning remains largely unexplored.

To tackle these issues, we first design the task-specific KG structures, to transform general graphs and their tasks into KG formats. As shown in Figure~\ref{fig:3}, we introduce two new entities, ``label $\square$'', and ``super graph $\triangle$'', and define three new relations, ``node $\bigcirc$ is attributed with label $\square$'', ``node $\bigcirc$ belongs to super graph $\triangle$'', ``super graph $\triangle$ is attributed with label $\square$''. These definitions allow us to reframe classification tasks as KG reasoning, predicting the tail entity based on a given relation and head entity. For example, in a citation network, performing node classification is analogous to reasoning edges between papers and labels. 

Existing KG reasoning models mainly capture invariant structural patterns while overlooking semantic features, resulting in the semantic isolation issue. Simply adding semantic features at initiation does not effectively balance topological generalization with domain-specific semantics. Instead, we propose a novel Semantic Conditional Message Passing (SCMP) that incorporates both local semantic neighbors and global semantic information, to enhance the performance of KG reasoning while preserving topological generalizability.

In summary, we propose the Semantic Conditional Reasoner (\method), a novel graph reasoning framework that leverages inductive KG reasoning to overcome semantic isolation and advance graph foundation models. To the best of our knowledge, this is the first work to employ KG reasoning as a pretraining objective for graph foundation models.  
We conducted extensive experiments on link prediction, node classification, and graph classification tasks across 38 datasets from diverse domains. The results demonstrate performance improvement over existing foundation models and supervised baselines, underscoring the transferability of our approach.

\section{Related Works}


Transferability is key to the success of graph foundation models. Here we describe current studies by clarifying how they address differences between source and target datasets and bridge gaps across task formats.
First, OFA~\citep{OFA} and ZeroG~\citep{zerog} leveraged pre-trained language models to encode node/class features as text, creating a unified feature space across diverse datasets. 
Meanwhile, OpenGraph~\citep{OpenGraph} adopted masked autoencoding, and AnyGraph~\citep{AnyGraph} used link prediction loss during pretraining on multiple graphs, enabling direct application to conduct node classification and link prediction tasks on new graphs. 

Beyond designing model architectures and pretraining objectives, graph prompt learning is also popular to employ lightweight prompts, aiming to align pre-training with downstream tasks~\citep{ProG}. 
Current notable approaches include GPPT~\citep{sun2022gppt}, All-in-one~\citep{sun2023all}, GPrompt~\citep{gong2023prompt}, and GPF-plus~\citep{gpfplus2023}. 

The existing foundation models in KGs study the transferability across different KGs~\citep{KG-ICL,TRIX}. For example, ULTRA~\citep{ultra} learned transferable graph representations by conditioning on relational interactions, enabling generalization to unseen KGs. 

We provide a detailed version of related works in Appendix~\ref{sec:relatedwork}. Unlike all the prior studies, our method~\method explores a novel scenario, training solely on common-sense KG datasets while achieving transferability across a wide range of general graphs and tasks without the need for extra fine-tuning.

\section{Background}



Let a knowledge graph be represented as $\graph = \{\ents, \rels, \triples\}$, where $\ents$ is the set of entities and $\rels$ is the set of relations. The factual triples in the KG are denoted by $\triples = \{(e_h, r, e_t) \mid e_h, e_t \in \ents, r \in \rels\}$, where each triple consists of a head entity $e_h$, a relation $r$, and a tail entity $e_t$.
Given a query $(e_q, r_q)$, where $e_q \in \ents$ is the query entity and $r_q \in \rels$ is the query relation, the goal of \textit{KG Reasoning} is to identify the correct entity $e_v \in \ents$, such that either $(e_q, r_q, e_v)$ or $(e_v, r_q, e_q)$ forms a valid triple in $\graph$.
In addition, we define a feature matrix $\features \in \mathbb{R}^{|\ents| \times d_0}$, where each row represents a feature vector of dimension $d_0$ for the corresponding entity in the set $\ents$. 
Now, consider a model trained on a knowledge graph $\gtrain = \{\etrain, \rtrain, \ttrain\}$. The task of zero-shot inductive reasoning on knowledge graphs is to test the model on a new inference graph $\ginf = \{\einf, \rinf, \tinf\}$, where both entities and relations are completely unseen during training. 
The whole notation used are listed 
in Appendix~\ref{app_sec:notations}.

\textbf{CMP-based Backbone Model:}
For inductive KG reasoning, recent studies utilize graph neural networks based on \emph{Conditional Message Passing} (CMP) to represent KG triples~\citep{NBF-NIPS21,REDGNN-WWW22,Adaprop,ANet,OurGraPE}.
Traditional message passing neural networks, such as GCN~\citep{GCN-ICLR17}, GAT~\citep{GAT-ICLR18}, and GraphSAGE~\citep{GraphSAGE}, compute \emph{unary} node representations and lack the ability to model interactions in a node set (such as edges)~\citep{RevisitGNN}.
Differently, for a KG $\graph$ and trainable relation embeddings $\mathbf{R}$, a CMP-based model $\mathcal{M}_\theta={\it CMP}($\textbf{q}$, \graph, \mathbf{R})$ calculates the triple representations $\vh_{v}$ for each entity $e_v$ conditioned on the query $\textbf{q} = (e_q, r_q)$:
\begin{align}
\begin{split}
&\vh_{v}^{(0)} = \init(e_q, e_v,\vr_q) = \mathbbm{1}_{e_q = e_v}  * \vr_q,
\end{split}
\label{eq:init1}
\\
\begin{split}
&\tilde{\vh}_{v}^{(l)} = \aggregate(\{\!\! \{ \mes(\vh_{w}^{(l)},\vr)|~  e_w \in \mathcal{N}_r(e_v), r \in \rels \}\!\!\}),
\end{split}
\\
\begin{split}
&\vh_{v}^{(l+1)} = \updatefunc(\vh_{v}^{(l)}, \tilde{\vh}_{v}^{(l)}).
\end{split}
\end{align} 
Here $\vh_{v}$ represents the aggregation of all paths connecting $e_q$ and $e_v$ based on $r_q$. Therefore, $\init()$ function only initializes the source entity $e_q$ by setting $\vh_{q}^{(0)}$ to the learnable relation vector $\vr_q=\mathbf{R}[r_q]$, while initializing all other entities to zero.
$\mes()$ is a differentiable message function that integrates two types of information: the aggregated paths between $e_q$ and $e_w$ as $\vh_{w}^{(l)}$, and the edge connecting $e_w$ to $e_v$ as $\vr=\mathbf{R}[r]$.
The representations are iteratively updated over $L$ layers through $\aggregate()$ and $\updatefunc()$ functions. The final representation $\vh_{v}^{(L)}$ is then used to predict the plausibility of triples $(e_q, r_q, e_v)$ in KG reasoning.



These conditional representations are theoretically expressive~\citep{LPTheory-NIPS23} and practically effective~\citep{ultra}. Using a specific initialization function $\init()$, CMP-based models rely solely on KG structures and relation embeddings, enabling inductive reasoning on new KGs. Moreover, CMP supports parallel learning of $\vh_{v}$ for all $e_v \in \ents$, reducing computational costs.

\begin{figure*}[!tb]
\centering
\includegraphics[width=1.0\textwidth]{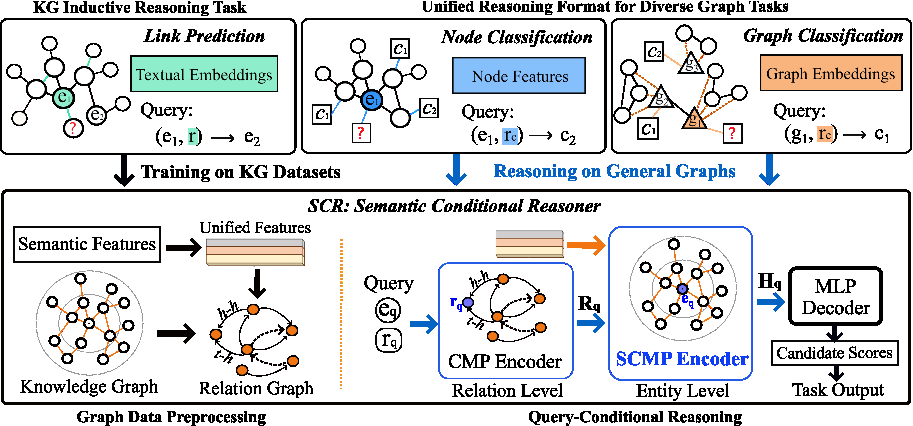}
\vspace{-4mm}
\caption{The proposed framework \method transforms diverse graph tasks into inductive reasoning on knowledge graphs with semantic features.}
\label{fig:3}
\vspace{-4mm}
\end{figure*} 

\section{Methodology}

We first define a unified graph reasoning format to achieve \textbf{cross-task transferability} in Sec.~\ref{sec:3.1}, where node classification and graph classification are reformulated as inductive reasoning tasks on a specific KG structure.
For \textbf{semantic transferability}, we then design the \emph{Semantic Conditional Message Passing} (SCMP) in Sec.~\ref{sec:3.3}, which enhances CMP's ability to effectively leverage semantic features while maintaining topological expressive power.
In Sec.~\ref{sec:3.2}, we illustrate the whole process of \method to handle unseen KGs with arbitrary entity and relation type. 
The overall framework of \method is present in Figure~\ref{fig:3}. 

\subsection{Unified Graph Reasoning Format}   
\label{sec:3.1}

Here we aim to develop a unified framework that addresses node-level, edge-level, and graph-level tasks simultaneously. First, KG reasoning can be considered a specialized form of link prediction focused on a specific relation, making it straightforward to apply. Because node and graph classification tasks draw labels from a finite set, we reformulate them using the following task-specific KG structure (Definition~\ref{def:task}), thereby transforming labeling tasks into KG reasoning.

\begin{definition} (Task-specific KG Structure) \label{def:task}
For a given graph task on a dataset $\mathcal{D}=(X, Y)$, the task-specific knowledge graph $\widetilde{G}$ is constructed as follows:
$$\widetilde{G} =\{(x_i,\text{is\_attributed\_with},y_i) | (x_i,y_i) \in \mathcal{D}\} \cup \mathcal{T}_X,$$
where $\mathcal{T}_X$ includes all original edges present within $X$.
\end{definition}

The examples are illustrated in Figure~\ref{fig:3}. In node classification tasks, a unique ``label $\square$'' entity is introduced for each label type, with a defined relation ``node $\bigcirc$ is attributed with label $\square$'' connecting nodes to their corresponding labels. The original node connections are preserved, and the input node features are also retained as semantic features for entities. For graph classification, we integrate individual graphs into a KG structure by adding ``super graph $\triangle$'' entities linked to their nodes via the relation ``node $\bigcirc$ belongs to super graph $\triangle$''. We then aggregate semantic features for each graph entity and add "semantically-nearest" edges between super graph nodes. The detailed procedures for task-specific KGs are given in Appendix~\ref{app:proof1}.

Therefore, an ideal, fully trained CMP-based graph model that generalizes across various knowledge graphs can simultaneously handle node and graph classification tasks.
Knowledge graphs often contain numerous “n-to-1” relations, such as “person-to-gender” or “movie-to-genre,” which are closely related to labeling tasks~\citep{ConvE}. This relationship enables KG reasoning models to achieve strong performance in the new relation “is\_attributed\_with”. Furthermore, this unified KG reasoning format eliminates the need to learn separate parameters for each label class, enabling support for unseen labels during inference.

\subsection{Semantic Conditional Message Passing (SCMP)}
\label{sec:3.3}

\begin{wrapfigure}{R}{0.5\textwidth}
\vspace{-4mm}
\centering
\setlength{\abovecaptionskip}{0cm} 
\setlength{\belowcaptionskip}{0cm}
\includegraphics[width=1\linewidth]{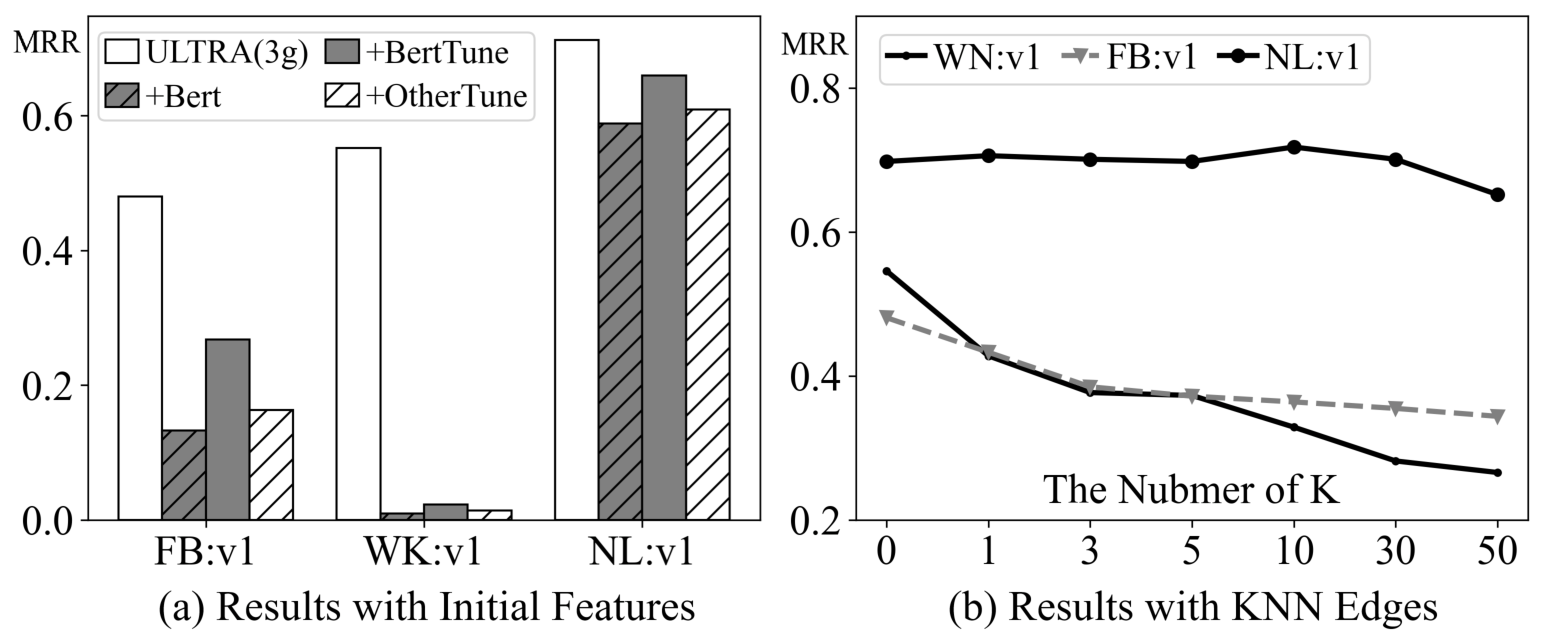}
\caption{Preliminary results of the baseline ULTRA. ``+Bert'' denotes using BERT-encoded features as initialization. We also fine-tune the BERT-encoded ULTRA (``+BertTune''), followed by reasoning with features from the other source (``+OtherTune''). Higher MRR is better.}
\label{fig:2}
\vspace{-4mm}
\end{wrapfigure}

\textbf{Semantic Isolation Issue:}
Due to the specific design for topological generalization, CMP-based models cannot effectively utilize node semantic features in general graph tasks, named as the semantic isolation issue.

Some simple attempts even worsen performance.
In the first attempt, we directly apply node semantic features into the $\init()$ function (Eq.~\ref{eq:init1}) to initialize node representation $\vh_{v}^{(0)}$, similar to standard GNNs. 
However, as shown in Figure~\ref{fig:2}(a), the performance degrades after injecting features (``+Bert'') and even after fine-tuning with such initialization (``+BertTune''). 
This is because the core \emph{target node distinguishability} assumption of CMP is violated~\citep{LPTheory-NIPS23}, which requires, for all $r_q \in \rels$ and $ e_v \neq e_q \in \ents$, the condition $\init(e_q,e_q,\vr_q) \neq \init(e_q,e_v,\vr_q)$ must hold~\citep{RevisitGNN}. 
Similar declines are observed when embeddings from other language models are used (``+OtherTune'').

The second attempt is to introduce semantic feature similarity into the graph structure. 
We construct a $k$-nearest neighbor (KNN) graph based on the similarity of node features, and add new triples with the relation "is\_semantic\_similar\_to" into the knowledge graph. 
As illustrated in Figure~\ref{fig:2}(b), increasing $k$ reduces the performance of ULTRA. This is due to the added edges diluting local information and causing distant nodes to lose their distinctiveness. The resulting dense topology either amplifies or compresses embeddings unevenly, leading to over-smoothing and a decline in link prediction performance.

Although these attempts fail, they offer insights for improving CMP from three aspects. First, we adapt a semantic unifier to preprocess node features. We then design the new semantic-injected INIT function satisfying the assumption of target node distinguishability. Finally, we use the parameter-frozen CMP to embed the semantic node features directly.

\textbf{Semantic Feature Unifier:}
To handle semantic diversity across domains, the semantic unifier is employed to preprocess node features without additional training.
Given the feature matrix $\features\in \mathbb{R}^{|\ents|\times d_0}$, we utilize singular value decomposition (SVD) in extracting important latent features:
\begin{align}
\begin{split}
\tilde{\features}= \text{LayerNorm}(\bm{U}\sqrt{\Lambda}), \ (\bm{U}, \Lambda, \bm{U}) = \text{SVD}(\features, d)
\end{split}
\end{align}
where $\text{LayerNorm}(\cdot)$ represents the layer normalization function, ensuring numerical stability. If $min(d_0, |\ents|)$ is smaller than $d$, SVD will use a reduced rank to decompose $\features$, with the remaining dimensions zero-padded to reach $d$. Such that, the unified features $\tilde{\features} \in \mathbb{R}^{|\ents|\times d}$ maintain consistent dimensionality $d$ across different graph data. Besides, the relative spatial distances between nodes are preserved in the unified features due to the nature of SVD. 
For scalability, we employ randomized truncated SVD to ensure linear complexity.

\textbf{Semantic-injected INIT Function:}
Given a query $\mathbf{q}=(e_q, \vr_q)$, we first recall the initialization function in CMP: $\init(e_q,e_v,\vr_q) = \mathbbm{1}_{e_q = e_v} * \vr_q$.
Instead of using the original semantic features, we inject the semantic neighbor labels into the entity initialization.
The improved initialization function is defined as follows:
\begin{align}
&\init^2(e_q,e_v,\vr_q) = \mathbbm{1}_{e_q = e_v} * \vr_q + \mathbbm{1}_{e_v \in \mathcal{S}_{e_q}} * \bm{v}_a,
\end{align}
where $\mathcal{S}_{e_q}$ represents the semantic neighbors of $e_q$. These neighbors are determined by selecting the top $k$ spatially nearest entities in the semantic space based on pairwise similarities, while excluding direct topological neighbors. In addition, $\bm{v_a}$ denotes a trainable vector shared for all semantic neighbors in $\mathcal{S}_{e_q}$.
In this schema, the initial representations of these neighbor entities are not all-zero vectors, enabling them to propagate effective high-order messages at the beginning of the CMP process.
Note that, according to the theoretical study of CMP~\citep{LPTheory-NIPS23}, if we assume $\bm{r}_q \neq \bm{v}_a$ and neither of them contains zero entries (Appendix \ref{app:Assumption} shows the assumption generally holds.), $\init^2$ function satisfies the \emph{target node distinguishability} assumption. Specifically, for all $r_q \in \rels$ and for any $e_v \neq e_q \in \ents$, it holds that $\init^2(e_q, e_q, \vr_q) \neq \init^2(e_q, e_v, \vr_q)$.

\textbf{Non-parametric Semantic Representation:}
Although the new initialization function captures high-level semantic relationships among entities, the original semantic features remain excluded from the computation process. 
To address this, we still use semantic features to initialize all the entities, but keep CMP parameters frozen. This setup is similar to SGC~\citep{wu2019simplifying}, a non-parametric adaptation of traditional GNNs that relies on repeated graph propagation for representation learning. Finally, an MLP is employed to merge the semantic representation $\mathbf{H}_g$ with the original CMP-based representations based on the specific query. 
\begin{align}
\begin{split}
&\mathbf{H}_g = {\it CMP}_{\theta}(\emptyset, \graph, \mathbf{R}_g) \quad \text{where}  \quad \mathbf{H}^{(0)}_g = \tilde{\features}
\label{eq:non-parametic_semantic_rep}
\end{split}
\\
\begin{split}
&\mathbf{H} = {\it CMP}_{\theta}(\textbf{q}, \graph, \mathbf{R}_q)+\text{MLP}(\mathbf{H}_g), 
\end{split}
\end{align}
where the parameters of both CMP instances are shared. 
The empty set used in Eq.~\ref{eq:non-parametic_semantic_rep} means that this CMP is query-independent, and we do not need to input the query. Besides, $\mathbf{R}_q$ and $\mathbf{R}_g$ are two parts of relation representations, which will be described in Sec.~\ref{sec:3.2}.
Notably, $\mathbf{H}_g$ can be precomputed and seamlessly integrated into the computation process of query-specific CMP, enabling SCMP to preserve time and space complexities compared to CMP.

\subsection{The Whole Process of \method}   %
\label{sec:3.2}

Here we describe the entire process of training \method on multiple KG datasets. As illustrated in the lower part of Figure~\ref{fig:3}, the first step is to preprocess the graph data. For each KG $\graph$, we apply the semantic feature unifier to process node features, followed by the construction of the relation graph $\graph_{\it r}$. The second step is query-conditional reasoning. Given a query $\mathbf{q}=(e_q, r_q)$ on $\graph$, we first apply CMP to learn the relation representation $\mathbf{R_q}$ via the relation graph $\graph_{\it r}$. Based on the relation representations $\mathbf{R_q}$, we further use our proposed SCMP to learn the triple representation $\vh_{v}$ for $(e_q, r_q, e_v)$. These representations are passed through an MLP to compute the scores for the existence of triples. We use cross-entropy loss to train the model for classifying positive and negative triples.

As for the inference phase, \method unifies classification tasks as KG inductive reasoning by transforming a general graph into a KG. 
Thereby, the learned reasoning patterns in \method can be adapted to label the entity without fine-tuning corresponding samples.


\textbf{Build the Relation Graph $\graph_{\it r}$:} Given a KG $\graph$, a relation graph $\graph_{\it r}$ is constructed following ULTRA~\citep{ultra}, to connect unseen relation types in $\graph$ with four types of relation-level interactions (i.e., "head-to-head", "tail-to-tail", "head-to-tail", and "tail-to-head"). Please refer to Appendix \ref{app:sarg} for further details.


\textbf{Learn the Relation Representation $\mathbf{R_q}$:} 
We then learn the relation embeddings via $\graph_{\it r}$. Specifically, given a query $\mathbf{q}=(e_q, r_q)$, the calculation process is as follows:
\begin{align}
\begin{split}
&\mathbf{R}_g = {\it CMP}_{\phi}(\emptyset, \graph_r, \mathbf{P}) \quad \text{where}  \quad \mathbf{R}^{(0)}_g = \mathbf{1}
\end{split}
\\
\begin{split}
&\mathbf{R}_q={\it CMP}_{\phi}(\mathbf{q}, \graph_{\it r}, \mathbf{P})
\end{split}
\end{align}
where $\mathbf{P}$ denotes the learnable embeddings corresponding to four types of interactions in the relation graph $\graph_{\it r}$. 
For the query $\mathbf{q}$, the query-conditional relation representations $\mathbf{R}_q$ are generated using CMP on $\graph_{\it r}$. Alternatively, when no query is provided and the initialized embedding $\mathbf{R}_g^{(0)}$ is set as an all-ones vector, the query-independent representations $\mathbf{R}_g$ are computed and utilized in Eq. \ref{eq:non-parametic_semantic_rep}.

\textbf{Query Conditional Reasoning:} 
Based on $\mathbf{R_q}$ and $\mathbf{R}_g$, we utilize our proposed SCMP model to learn the entity representation given the query $\mathbf{q}=(e_q, r_q)$:
\begin{align}
\begin{split}
&\mathbf{H}={\it SCMP}_{\theta}(\mathbf{q}, \graph, \tilde{\features}, \mathbf{R}_q, \mathbf{R}_g), 
\end{split}
\label{eq:glb-com} 
\\
\begin{split}
&p(\bm{q}, e_v) = \text{MLP}(\vh_{v})
\end{split}
\end{align}  
where $\vh_{v} \in \bm{H}$ denotes the final entity representation of the entity $e_v$.
To evaluate the plausibility of the triple $(e_q, r_q, e_v)$, an MLP is employed to compute a score, where a higher value indicates a greater likelihood of the triple being valid in $\graph$. 

\textbf{Model Training:} 
KG inductive reasoning models are typically trained by minimizing the binary cross-entropy loss over positive and negative triples. 
To handle semantic features across domains, we train one \method model with multiple types of semantic features on diverse KG datasets. 
Specifically, we employ the BERT~\citep{Bert} sentence encoder to generate semantic features. We also incorporate ontology features and explore a non-feature scenario during model training, as presented in Appendix~\ref{app:imd}. 
Note that, we focus on node semantics in this work, edge semantics can be supported with trivial modifications to our framework. 
At predefined mini-batch intervals, the feature type is reselected to help the model adapt to diverse input features and improve its generalization ability. 
The total pretraining loss is computed as follows:
\begin{align}
\begin{split}
    &\mathcal{L} = \sum_{\features\in\mathcal{F}} \left(- \log p(\bm{q}, e_a | \features) - \frac{1}{n}\sum_{i=1}^{n}  \log (1 - p(\bm{q}, e_i | \features)) \right)\nonumber
\end{split}
\end{align}
Here, $p(\bm{q}, e_a | \features)$ is the score for a positive triple in KG $\graph$ with the node features $\features$, while $\{(\bm{q}, e_i)| \features\}^n_{i=1}$ contains negative samples created by corrupting the target entity.

We analyze the expressive power of our proposed \method in Appendix~\ref{app:proof}, and discuss the computational complexity and scalability in Appendix \ref{sec:complexity}. 

\section{Experiments}
\label{sec:4}
We evaluate our method on 38 diverse datasets across three-level tasks. In particular, we wish to answer the following research questions: 
\textbf{RQ1}: How effective is \method in inductive reasoning across distinct knowledge graphs?
\textbf{RQ2}: To what extent does \method generalize across diverse feature spaces on the same KG? 
\textbf{RQ3}: How well does \method generalize across a variety of graph-related tasks? 
\textbf{RQ4}: What is the impact of the main components on the performance of \method?
\textbf{RQ5}: How does the reasoning performance change when adjusting the key hyperparameters?
Due to the space limitation, discussions about RQ5 are detailed in Appendix~\ref{exp:RQ5}.

\begin{table*}[!t]
\caption{Performance on KG inductive reasoning datasets. ``(3g)'' means training with three KGs, and ``\method-X'' refers to results obtained using different types of semantic features (e.g., ``One'' means all-ones features). 
The best results are in bold. }
\vspace{-2mm}
\label{tab:linkmain}
\setlength\tabcolsep{2pt}
\scriptsize
\begin{tabular}{lcccccccccccccc}
\toprule
\multirow{2}{*}{Methods} & \multicolumn{2}{c}{IndE(FB)} & \multicolumn{2}{c}{IndE(WN)} & \multicolumn{2}{c}{IndE(NL)} & \multicolumn{2}{c}{IndER(FB)} & \multicolumn{2}{c}{IndER(WK)} & \multicolumn{2}{c}{IndER(NL)} & \multicolumn{2}{c}{\textbf{Total AVG}}\\
 & MRR & Hits@10 & MRR & Hits@10 & MRR & Hits@10 & MRR & Hits@10 & MRR & Hits@10 & MRR & Hits@10 & MRR & Hits@10 \\
\midrule
Supervised SOTA & 0.477 & 0.636 & 0.640 & 0.734  & 0.464 & 0.654 & 0.166 & 0.296 & 0.152 & 0.244 & 0.296 & 0.481 & 0.366	& 0.507\\
  \midrule
ULTRA(3g) & 0.486 & 0.667 & 0.517 & 0.678 & 0.561 & 0.742 & 0.386 & 0.599 & \textbf{0.254} & 0.403 & 0.393 & 0.561 & 0.433	& 0.608\\
ULTRA(4g) &	0.491 &	0.670 &	0.567 &	0.689 &	\textbf{0.616} &	\textbf{0.803} &	0.387 &	0.598 &	0.251 &	\textbf{0.415} &	0.398 &	0.588 &	\textbf{0.451} &	0.627\\
ULTRA(50g) &	0.493 &	0.664 &	0.558 &	0.664 &	0.590 &	0.777	 & 0.382 &	0.585 &	0.251 & 0.406 &	0.397 &	0.582 &	0.445	& 0.613\\
ProLINK(3g) & 0.494 &	0.684 &	0.553 &	0.690 &	0.546 &	0.759 &	0.372 &	0.591 &	0.234 &	0.393 &	0.400 &	0.590 &	0.433 &	0.618 \\
 \midrule
\textbf{\method(3g)} & \textbf{0.495} & \textbf{0.688} & \textbf{0.576} & \textbf{0.703}  & 0.592 & 0.791 & \textbf{0.392} & \textbf{0.611} & 0.251 & 0.407 & \textbf{0.403} & \textbf{0.599} & \textbf{0.451} &	\textbf{0.633}\\
\midrule
\midrule
\method-One & 0.491 & 0.678 & 0.569 & 0.688  & 0.581 & 0.773 & 0.390 & 0.604 & 0.250 & 0.399 & 0.388 & 0.578 & 0.445 &	0.620\\
\method-MPNet & 0.495 & 0.688 & 0.578 & 0.704  & 0.589 & 0.788 & 0.392 & 0.611 & 0.250 & 0.406 & 0.403 & 0.601 & 0.451 &	0.633\\
\method-MiniLM & 0.496 & 0.687 & 0.576 & 0.702 & 0.585 & 0.788 & 0.392 & 0.611 & 0.250 & 0.406 & 0.405 & 0.604 & 0.451	& 0.633\\
\method-DistilBert & 0.495 & 0.688 & 0.576 & 0.706  & 0.584 & 0.788 & 0.392 & 0.610 & 0.250 & 0.407 & 0.401 & 0.601 & 0.450	& 0.633\\
\method-Ontology & 0.489 & 0.684 & 0.570 & 0.679 & 0.575 & 0.772 & 0.387 & 0.605 & 0.230 & 0.395 & 0.391 & 0.584 & 0.440	& 0.620\\
\bottomrule
\end{tabular}
\vspace{-6mm}
\end{table*}

\subsection{Experimental Setup}

We pre-train \method on three commonly-used KG datasets, WN18RR \citep{WN18RR}, FB15k237 \citep{FB15k237}, and CodexM~\citep{Codex}.
The CMP follows NBFNet with a non-parametric DistMult~\citep{DistMult} message function and a simplified PNA aggregation function~\citep{OurGraPE}. 
For semantic features, we employ the BERT~\citep{Bert} sentence encoder to generate pre-training features. 
Hyperparameters are selected through grid search based on the metrics from the validation set without fine-tuning for each dataset.
Implementation details and hyperparameter configurations are provided in Appendix~\ref{app:imd}.
Three graph learning tasks are used to evaluate: link-level KG inductive reasoning and node-/graph-level classification on general graphs, across 38 real-world datasets.
The details of tasks and datasets are described in Appendix~\ref{app:data}.

\subsection{Main Experimental Results (RQ1)}

All 24 inductive KG reasoning datasets are used in RQ1: 
the first 12 datasets from GraIL~\citep{GraIL-ICML19} with test graphs containing only unseen entities (termed as ``IndE''), and the remaining 12 datasets from InGram~\citep{ingram} featuring both unseen entities and relations (termed as ``IndER''). Notably, eight datasets in IndE/IndER(NL) come from NELL-995 (excluded from training), introducing new semantic features for each method.
This setting prevents data leakage by dynamically generating entity representations based on the unique structure of each KG during training and inference. Even if a triple appears in both the pre-training and test datasets, the different structures around it ensure distinct representations, thereby mitigating memorization.

We compare \method with two KG reasoning baselines (ULTRA and ProLINK pre-trained on different sizes of KGs) and one supervised SOTA. 
Two evaluation metrics are used: Mean Reciprocal Rank (MRR) and Hits@N, where higher scores indicate better performance~\citep{TransE,GraIL-ICML19}.
For semantic features, we generate entity-level embeddings from available textual descriptions using BERT sentence encoders.

We report the average performance for each benchmark and the results are summarized in Table~\ref{tab:linkmain}. 
A comprehensive evaluation on more than 50 transductive and inductive KG datasets is provided in Appendix \ref{app:50KGs}.
Overall, \method outperforms all existing zero-shot models as well as the supervised model in the total average metrics, demonstrating its effectiveness. 
For individual benchmarks, we observe that \method surpasses ULTRA(3g), ProLINK, and supervised results in most metrics. 
Although ULTRA(4g) and ULTRA(50g) achieve better performance on some metrics, they are pre-trained on more diverse KGs (including NELL-995) and still show poorer performance on the IndER(FB) and IndE(WN) benchmarks. 
Furthermore, compared to IndER(X) benchmarks, \method shows substantial performance gains on IndE(X) benchmarks beneficial from node semantic features.

\begin{table*}[!t]
\centering
\setlength\tabcolsep{3pt}
\caption{The accuracy results on node classification datasets. 
GraphAny(X) or AnyGraph(X) indicates pertaining on the X dataset. 
\method-20\% uses 20\% of the ``node-label'' edges from the training set, while \method-5 includes five edges per class.
The best results are bolded.}
\vspace{-2mm}
\small
\label{tab:node}
\begin{adjustbox}{max width=\textwidth}
\begin{tabular}{llccccccc}
\toprule
Learning Paradigm & Methods & Cora & Citeseer & Pubmed & Wisconsin & Texas & Actor &  \textbf{Avg.Rank}\\
\midrule
\multirow{3}{*}{Full-Shot Training} & MLP & 48.42\scriptsize{$\pm$0.63} & 44.40\scriptsize{$\pm$0.44} & 69.50\scriptsize{$\pm$1.79} & 66.67\scriptsize{$\pm$3.51} & 48.65\scriptsize{$\pm$4.01} & 33.95\scriptsize{$\pm$0.80} & 8.83 \\
& GCN & 81.40\scriptsize{$\pm$0.70} & 63.40\scriptsize{$\pm$0.63} & 76.60\scriptsize{$\pm$0.32} & 37.25\scriptsize{$\pm$1.64} & 51.35\scriptsize{$\pm$2.71} & 28.55\scriptsize{$\pm$0.68} & 7.33 \\
& GAT & 81.70\scriptsize{$\pm$1.43} & 69.10\scriptsize{$\pm$1.59} & 77.30\scriptsize{$\pm$0.60} & 52.94\scriptsize{$\pm$3.10} & 54.05\scriptsize{$\pm$2.41} & 27.30\scriptsize{$\pm$0.22} & 5.67 \\
\midrule
\multirow{4}{*}{\makecell{Graph Pre-Training\\Full-Shot Analytical\\Tuning}} & GraphAny(Products) & 79.36\scriptsize{$\pm$0.23} & 67.94\scriptsize{$\pm$0.29} & 76.54\scriptsize{$\pm$0.34} & 65.89\scriptsize{$\pm$2.23} & \textbf{73.52\scriptsize{$\pm$2.96}} & 28.99\scriptsize{$\pm$0.61} & 5.00\\
& GraphAny(Arxiv) & 79.38\scriptsize{$\pm$0.16} & 68.34\scriptsize{$\pm$0.23} & 76.36\scriptsize{$\pm$0.17} & 65.10\scriptsize{$\pm$3.22} & 72.97\scriptsize{$\pm$2.71} & 28.60\scriptsize{$\pm$0.21} & 5.50 \\
& GraphAny(Wisconsin) & 77.82\scriptsize{$\pm$1.15} & 67.50\scriptsize{$\pm$0.44} & 77.46\scriptsize{$\pm$0.30} & \textbf{71.77\scriptsize{$\pm$5.98}} & 73.51\scriptsize{$\pm$1.21} & \textbf{29.51\scriptsize{$\pm$0.55}} & 4.33\\
& GraphAny(Cora) & 80.18\scriptsize{$\pm$0.13} & 68.90\scriptsize{$\pm$0.07} & 76.60\scriptsize{$\pm$0.31} & 61.18\scriptsize{$\pm$5.08} & 71.89\scriptsize{$\pm$1.48} & 27.91\scriptsize{$\pm$0.16} & 5.17\\
\midrule
\multirow{3}{*}{\makecell{Graph Pre-Training\\No Tuning}} & OpenGraph & 80.65\scriptsize{$\pm$0.69}	& 69.99\scriptsize{$\pm$0.83} &	80.15\scriptsize{$\pm$1.28}	 & 24.42\scriptsize{$\pm$5.64}	& 21.78\scriptsize{$\pm$6.32}	& 16.74\scriptsize{$\pm$5.68} & 7.33\\
& AnyGraph(Link1) & 58.57\scriptsize{$\pm$7.82}	& 51.93\scriptsize{$\pm$6.04}	& 62.75\scriptsize{$\pm$2.55}	& 1.51\scriptsize{$\pm$0.37}	& 0.57\scriptsize{$\pm$0.19}	& 5.49\scriptsize{$\pm$0.31} & 12.83\\
& AnyGraph(Link2) & 69.05\scriptsize{$\pm$4.71}	& 45.52\scriptsize{$\pm$4.26}	& 78.02\scriptsize{$\pm$1.46}	& 1.29\scriptsize{$\pm$0.24}	& 0.81\scriptsize{$\pm$0.60}	& 5.56\scriptsize{$\pm$0.21} & 11.00\\
\midrule
\multirow{2}{*}{\makecell{KG Pre-Training\\No Tuning}} & ULTRA(3g) & 79.40\scriptsize{$\pm$0.00} & 67.40\scriptsize{$\pm$0.00} &  77.90\scriptsize{$\pm$0.00} & 49.02\scriptsize{$\pm$0.00} & 56.76\scriptsize{$\pm$0.00} & 22.61\scriptsize{$\pm$0.00} & 7.50\\
& \textbf{\method(3g)} & \textbf{81.80\scriptsize{$\pm$1.02}} & \textbf{71.33\scriptsize{$\pm$0.27}} & \textbf{82.93\scriptsize{$\pm$0.55}} & 54.91\scriptsize{$\pm$1.51} & 67.64\scriptsize{$\pm$0.44} & 23.26\scriptsize{$\pm$0.56} & \textbf{3.67}\\
\midrule
\midrule
\multirow{2}{*}{\makecell{Few-Shot Labeling}} & \method-20\% & 73.62\scriptsize{$\pm$2.78} & 56.50\scriptsize{$\pm$3.53} & 71.94\scriptsize{$\pm$0.23} & 50.59\scriptsize{$\pm$3.14} & 64.32\scriptsize{$\pm$3.59} & 22.78\scriptsize{$\pm$0.49} & 9.00\\
& \method-5 & 54.48\scriptsize{$\pm$1.96} & 32.38\scriptsize{$\pm$3.58} & 50.38\scriptsize{$\pm$6.17} & 28.63\scriptsize{$\pm$8.82} & 58.38\scriptsize{$\pm$4.39} & 19.00\scriptsize{$\pm$1.22} & 11.67\\
\bottomrule
\end{tabular}
\end{adjustbox}
\vspace{-2mm}
\end{table*}

\begin{table*}[!t]
\centering
\setlength\tabcolsep{3pt}
\caption{The accuracy results on graph classification datasets. \method-20\% and \method-5 are two few-shot labeling variants of \method. The best results are bolded.
}
\small
\label{tab:graph}
\begin{adjustbox}{max width=\textwidth}
\begin{tabular}{llccccccccc}
\toprule
Learning Paradigm & Methods & IMDB-B & COLLAB & PROTEINS & MUTAG & ENZYMES & COX2 & BZR & DD  &  \textbf{Avg.Rank}\\
\midrule
\multirow{1}{*}{Full-Shot Training} & GIN & \textbf{67.75\scriptsize{$\pm$2.50}}  & 58.20\scriptsize{$\pm$10.22} & 64.72\scriptsize{$\pm$0.84} & 75.50\scriptsize{$\pm$5.74} & 21.88\scriptsize{$\pm$0.55} & 77.90\scriptsize{$\pm$1.57} & \textbf{81.79\scriptsize{$\pm$2.94}} & \textbf{70.59\scriptsize{$\pm$0.81}} & 3.13\\  
\midrule
\multirow{2}{*}{One-Shot Training} & GCN & 57.30\scriptsize{$\pm$0.98} & 47.23\scriptsize{$\pm$0.61} & 56.36\scriptsize{$\pm$7.97} & 65.20\scriptsize{$\pm$6.70} & 20.58\scriptsize{$\pm$2.00} & 27.08\scriptsize{$\pm$1.95} & 25.80\scriptsize{$\pm$6.53} & 55.33\scriptsize{$\pm$6.22} & 10.25\\
& Pretrain\&Finetune & 57.75\scriptsize{$\pm$1.22} & 48.10\scriptsize{$\pm$0.23} & 63.44\scriptsize{$\pm$3.64} & 65.47\scriptsize{$\pm$5.89} & 22.21\scriptsize{$\pm$2.79} & 76.19\scriptsize{$\pm$5.41} & 34.69\scriptsize{$\pm$8.50} & 57.15\scriptsize{$\pm$4.32} & 7.25 \\
\midrule
\multirow{5}{*}{\makecell{Graph Pre-Training\\One-Shot Tuning}} & GPPT & 50.15\scriptsize{$\pm$0.75} & 47.18\scriptsize{$\pm$5.93} & 60.92\scriptsize{$\pm$2.47} & 60.40\scriptsize{$\pm$15.43} & 21.29\scriptsize{$\pm$3.79} & \textbf{78.23\scriptsize{$\pm$1.38}} & 59.32\scriptsize{$\pm$11.22} & 57.69\scriptsize{$\pm$6.89} & 8.50\\
& All-in-one & 60.07\scriptsize{$\pm$4.81} & 51.66\scriptsize{$\pm$0.26} & 66.49\scriptsize{$\pm$6.26} & 79.87\scriptsize{$\pm$5.34} & \textbf{23.96\scriptsize{$\pm$1.45}} & 76.14\scriptsize{$\pm$5.51} & 79.20\scriptsize{$\pm$1.65} & 59.72\scriptsize{$\pm$1.52} & 3.75\\
& GPrompt & 54.75\scriptsize{$\pm$12.43} & 48.25\scriptsize{$\pm$13.64} & 59.17\scriptsize{$\pm$11.26} & 73.60\scriptsize{$\pm$4.76} & 22.29\scriptsize{$\pm$3.50} & 54.64\scriptsize{$\pm$9.94} & 55.43\scriptsize{$\pm$13.69} & 57.81\scriptsize{$\pm$2.68} & 7.50\\
& GPF & 59.65\scriptsize{$\pm$5.06} & 47.42\scriptsize{$\pm$11.22} & 63.91\scriptsize{$\pm$3.26} & 68.40\scriptsize{$\pm$5.09} & 22.00\scriptsize{$\pm$1.25} & 65.79\scriptsize{$\pm$17.72} & 71.67\scriptsize{$\pm$14.71} & 59.36\scriptsize{$\pm$1.18} & 6.25\\
& GPF-plus & 57.93\scriptsize{$\pm$1.62} & 47.24\scriptsize{$\pm$0.29} & 62.92\scriptsize{$\pm$2.78} & 65.20\scriptsize{$\pm$6.04} & 22.92\scriptsize{$\pm$1.64} & 33.78\scriptsize{$\pm$1.52} & 71.17\scriptsize{$\pm$14.92} & 57.62\scriptsize{$\pm$2.42} & 7.25\\
\midrule
\multirow{2}{*}{\makecell{KG Pre-Training\\No Tuning}} &ULTRA(3g) & 49.25\scriptsize{$\pm$0.00}	& 51.80\scriptsize{$\pm$0.00}	& 58.09\scriptsize{$\pm$0.00}	& 63.33\scriptsize{$\pm$0.00}	& 15.21\scriptsize{$\pm$0.00}	& 77.75\scriptsize{$\pm$0.00}	& 79.32\scriptsize{$\pm$0.00}	& 43.52\scriptsize{$\pm$0.00} & 8.50\\
& \textbf{\method(3g)} & 61.83\scriptsize{$\pm$1.60}	& \textbf{65.45\scriptsize{$\pm$1.05}}	& \textbf{68.54\scriptsize{$\pm$1.47}}	& \textbf{85.33\scriptsize{$\pm$2.11}}	& 22.92\scriptsize{$\pm$2.03}	& 78.08\scriptsize{$\pm$1.33}	& 79.32\scriptsize{$\pm$0.06}	& 69.96\scriptsize{$\pm$0.74} & \textbf{1.75}\\
\midrule
\midrule
\multirow{2}{*}{\makecell{Few-Shot Labeling}} & \method-20\% & 53.45\scriptsize{$\pm$3.46} & 60.72\scriptsize{$\pm$1.07} & 66.13\scriptsize{$\pm$4.08} & 52.93\scriptsize{$\pm$14.37} & 17.25\scriptsize{$\pm$1.29}
& 75.5\scriptsize{$\pm$5.06} & 79.51\scriptsize{$\pm$0.37} & 69.75\scriptsize{$\pm$3.19} & 6.13\\
& \method-5 & 53.37\scriptsize{$\pm$2.83} & 46.25\scriptsize{$\pm$8.93} & 61.69\scriptsize{$\pm$8.57} & 80.27\scriptsize{$\pm$5.82} & 22.58\scriptsize{$\pm$1.15} & 58.12\scriptsize{$\pm$2.11} & 46.05\scriptsize{$\pm$12.11} & 62.14\scriptsize{$\pm$4.5} & 7.38\\
\bottomrule
\end{tabular}
\end{adjustbox}
\vspace{-6mm}
\end{table*}

\subsection{Generalizing Across Semantic Spaces (RQ2)}

To address the cross-domain challenge, we explore the generalizability of \method across different semantic feature spaces. 
Specifically, we select five types of semantic features as input to the pre-trained \method (within BERT-encoded semantic space). In Table~\ref{tab:linkmain}, ``MPNet'', ``MiniLM'', and ``DistilBERT'' refer to three popular sentence encoders based on language models, while ``One'' and ``Ontology'' utilize all-ones features and ontology features derived from relation type counting, respectively.
The results show that \method variants using different sentence encoders achieve performance nearly identical to the original \method, despite the encoders producing embeddings with varying dimensions. This consistency underscores the effectiveness of our unified semantic space. 
Although \method(One) and \method(Ontology) perform slightly worse than \method on some metrics, they still outperform baselines such as ULTRA(3g). It indicates that \method is not constrained by access to textual features and exhibits generalization capabilities across diverse feature sources, even all-ones features.
The semantic content of IndE(NL) and IndER(NL) was not included in the pre-training KGs. Despite this, the performance improvement still demonstrate the robustness of \method to unseen domains or semantic inputs.

\subsection{Generalizing Across Graph Tasks (RQ3)}

In this section, we verify the performance of \method on classification tasks across diverse general graphs.
Following prior studies~\citep{ProG}, we employ six node classification datasets, including homophilic graphs (Cora, Citeseer, PubMed)~\citep{sen2008homoCoraCiteseer,namata2012homoPubmed} 
and heterophilic graphs (Wisconsin, Texas, Actor)~\citep{pei2019heterophilicDatasets,tang2009social}.
We utilize eight graph classification datasets from various domains, covering social networks (IMDB-B, COLLAB)~\citep{yanardag2015deep}, 
biological (ENZYMES, PROTEINS, DD)~\citep{dobson2003distinguishing,borgwardt2005protein,wang2022faith}, 
and small molecule datasets (MUTAG, COX2, BZR)~\citep{MUTAG, rossi2015network}. 
The experimental results of prediction accuracy are shown in Table \ref{tab:node} and Table \ref{tab:graph}. Comprehensive results can be found in Appendix \ref{app:addexp}.

For node-level tasks, \method significantly outperforms existing foundation models on three homophilic graphs, and shows superior performance over ULTRA, OpenGraph, and AnyGraph on three heterophilic graphs. 
However, \method lags behind GraphAny on heterophilic graphs, as GraphAny utilizes training labels to tune the model parameters through an analytical solution. Such task-specific tuning in GraphAny and graph prompt methods (see Table \ref{tab:node2} in Appendix \ref{app:addexp} for more results) is powerful but limits their versatility as graph foundation models.

Table \ref{tab:graph} shows the experimental results for graph-level tasks. Following the experimental setup from ProG~\citep{ProG}, 80\% of graph samples are divided into the test set and only a few labeled graphs are transformed into the task-specific KG. We observe that \method outperforms existing graph models using one-shot training or prompt tuning. Further, \method is competitive with the fully-trained GIN using the same training set, while GIN requires extra training time and \method directly infers on the task-specific KG.
Existing zero-shot foundation models such as OpenGraph, AnyGraph, and GraphAny are not suitable for graph classification tasks. This limitation underscores the value of \method across diverse graph tasks.

\subsection{Ablation Studies (RQ4)}
\begin{wrapfigure}{R}{0.5\textwidth}
\vspace{-4mm}
\centering
\setlength{\abovecaptionskip}{0cm} 
\setlength{\belowcaptionskip}{0cm}
\includegraphics[width=1\linewidth]{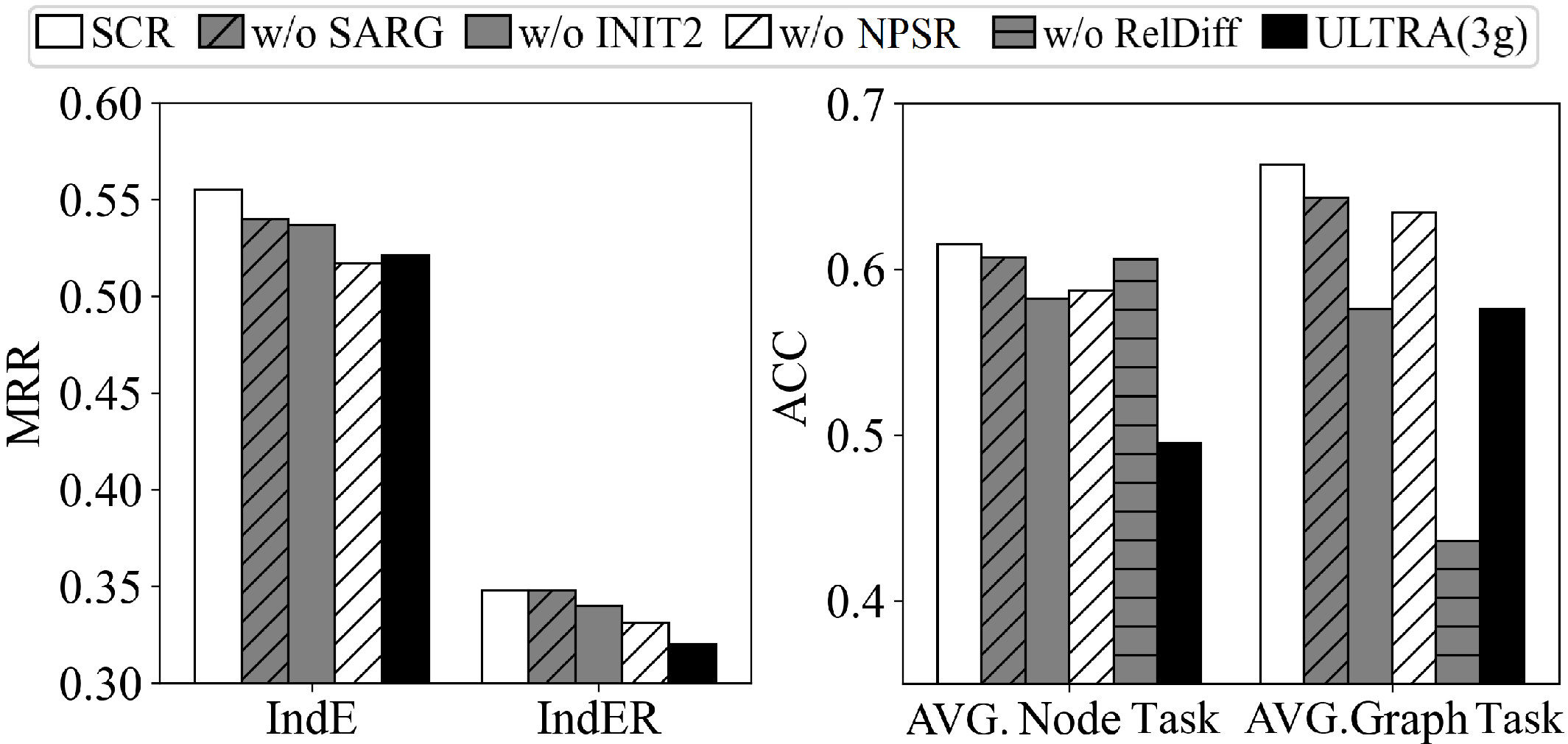}
\caption{The ablation study results of \method variants. ``w/o SARG'' denotes using the relation graph without semantic neighbors, the latter two have no semantic-injected initialization and global semantic encoding, respectively. ``w/o RelDiff'' denotes using identical relation embeddings in SCMP.}
\label{fig:ablation}
\vspace{-4mm}
\end{wrapfigure}

To assess the impact of the key techniques, we conduct ablation experiments for multiple pre-trained variants. The results are illustrated in Figure \ref{fig:ablation}.

\noindent\textbf{(1) Semantic Neighbors:} 
The two variants, excluding Semantic-Augmented Relation Graph (``w/o SARG'') and Semantic-Injected Entity Initialization (`w/o INIT2''), utilize the original relation graph and initialization function of ULTRA(3g) to omit semantic neighbor information. Across three tasks, the observed performance declines underscore the effectiveness of our semantic-injection techniques. The new INIT2 function has a relatively larger contribution, especially in graph-level tasks.

\noindent\textbf{(2) Non-parametric Semantic Representation:}
The non-parametric semantic encoding in SCMP directly handles semantic features, with the reduced performance of ``w/o NPSR'' emphasizing its essential role in leveraging semantic diversity. However, except for IndE datasets, the observation that ``w/o NPSR'' outperforms the original ULTRA in most tasks suggests that the post-merging approach effectively preserves the functionality of CMP.

\noindent\textbf{(3) Relational Condition:}
The variant ``w/o RelDiff'' utilizes the average vector of $\mathbf{R}_q$ as a substitute for each individual vector in $\mathbf{R}_q$, effectively removing the influence of relation type differences on entity-level inference. 
Despite the presence of only a few relation types in node/graph-level graph structures, the observed performance decline highlights the essential role of relational information. 
Notably, in graph-level tasks, relation-specific embeddings play a vital role in distinguishing rare graph-label relations from semantic and topological edges.

\noindent\textbf{(4) Few-Shot Labeling:}
We evaluate the performance of \method in scenarios where the 'node/graph-label' edges are sparse within the graph.
As shown in Table \ref{tab:node} and Table \ref{tab:graph},
reducing the number of labels in the task-specific KG leads to a decline in reasoning performance.
However, ``\method-20\%'' and ``\method-5'' still obtain better accuracy than some baselines, which highlights the effectiveness of our method.
In a few small-scale datasets, such as MUTAG and ENZYMES, the performance of ``\method-20\%'' is lower than ``\method-5'' due to the former having fewer accessible label edges.


\section{Conclusion}
\label{sec:con}
In this paper, we take the first successful step toward generalizing inductive KG reasoning in graph foundation models. 
The proposed method, \method, conducts semantic conditional message passing on multi-relational graphs, effectively integrating semantic features with structural information while addressing node-level, edge-level, and graph-level tasks simultaneously.
Extensive experiments demonstrate that \method achieves strong generalizability across diverse graph domains and tasks. We further discuss the limitations of our work in Appendix \ref{sec:limit}.
Our future research agenda will prioritize two key objectives: (1) extending the applicability of this approach to broader graph-based tasks across diverse domains, and (2) systematically validating the scaling principles that govern large-scale model architectures.

\bibliographystyle{icml2025}
\bibliography{bibtex}

\begin{thebibliography}{79}
\providecommand{\natexlab}[1]{#1}
\providecommand{\url}[1]{\texttt{#1}}
\expandafter\ifx\csname urlstyle\endcsname\relax
  \providecommand{\doi}[1]{doi: #1}\else
  \providecommand{\doi}{doi: \begingroup \urlstyle{rm}\Url}\fi

\bibitem[Bordes et~al.(2013)Bordes, Garc{\'{\i}}a{-}Dur{\'{a}}n, Weston, and Yakhnenko]{TransE}
Bordes, A., Garc{\'{\i}}a{-}Dur{\'{a}}n, A., Weston, J., and Yakhnenko, O.
\newblock Translating embeddings for modeling multi-relational data.
\newblock In \emph{Proceedings of the 27th Annual Conference on Neural Information Processing Systems, December 5-8, 2013, Lake Tahoe, Nevada, United States}, pp.\  2787--2795, 2013.

\bibitem[Bordes et~al.(2014)Bordes, Glorot, Weston, and Bengio]{WN18RR}
Bordes, A., Glorot, X., Weston, J., and Bengio, Y.
\newblock {A Semantic Matching Energy Function for Learning with Multi-relational Data}.
\newblock \emph{Machine Learning}, 94:\penalty0 233--259, 2014.

\bibitem[Borgwardt et~al.(2005)Borgwardt, Ong, Sch{\"o}nauer, Vishwanathan, Smola, and Kriegel]{borgwardt2005protein}
Borgwardt, K.~M., Ong, C.~S., Sch{\"o}nauer, S., Vishwanathan, S., Smola, A.~J., and Kriegel, H.-P.
\newblock Protein function prediction via graph kernels.
\newblock \emph{Bioinformatics}, 21\penalty0 (suppl\_1):\penalty0 i47--i56, 2005.

\bibitem[Chami et~al.(2020)Chami, Wolf, Juan, Sala, Ravi, and R{\'e}]{GoogleAttH}
Chami, I., Wolf, A., Juan, D.-C., Sala, F., Ravi, S., and R{\'e}, C.
\newblock Low-dimensional hyperbolic knowledge graph embeddings.
\newblock In \emph{Proceedings of the 58th Annual Meeting of the Association for Computational Linguistics, {ACL} 2020, Online, July 5-10, 2020}, pp.\  6901--6914, 2020.

\bibitem[Chen et~al.(2021)Chen, He, Wu, and Wang]{TACT-AAAI21}
Chen, J., He, H., Wu, F., and Wang, J.
\newblock Topology-aware correlations between relations for inductive link prediction in knowledge graphs.
\newblock In \emph{Thirty-Fifth {AAAI} Conference on Artificial Intelligence, {AAAI} 2021, Thirty-Third Conference on Innovative Applications of Artificial Intelligence, {IAAI} 2021, The Eleventh Symposium on Educational Advances in Artificial Intelligence, {EAAI} 2021, Virtual Event, February 2-9, 2021}, pp.\  6271--6278. {AAAI} Press, 2021.
\newblock URL \url{https://doi.org/10.1609/aaai.v35i7.16779}.

\bibitem[Chen et~al.(2023)Chen, Liu, Liu, Li, Mao, and Sun]{chen2023ultradp}
Chen, M., Liu, Z., Liu, C., Li, J., Mao, Q., and Sun, J.
\newblock {ULTRA}-{DP}: {Unifying} {Graph} {Pre}-training with {Multi}-task {Graph} {Dual} {Prompt}.
\newblock \emph{arXiv preprint}, 2023.

\bibitem[Corso et~al.(2020)Corso, Cavalleri, Beaini, Li{\`{o}}, and Velickovic]{PNA-NIPS20}
Corso, G., Cavalleri, L., Beaini, D., Li{\`{o}}, P., and Velickovic, P.
\newblock Principal neighbourhood aggregation for graph nets.
\newblock In \emph{Advances in Neural Information Processing Systems 33: Annual Conference on Neural Information Processing Systems 2020, NeurIPS 2020, December 6-12, 2020, virtual}, 2020.

\bibitem[Cui et~al.(2024)Cui, Sun, and Hu]{KG-ICL}
Cui, Y., Sun, Z., and Hu, W.
\newblock A prompt-based knowledge graph foundation model for universal in-context reasoning.
\newblock \emph{CoRR}, abs/2410.12288, 2024.
\newblock \doi{10.48550/ARXIV.2410.12288}.
\newblock URL \url{https://doi.org/10.48550/arXiv.2410.12288}.

\bibitem[Dettmers et~al.(2018)Dettmers, Minervini, Stenetorp, and Riedel]{ConvE}
Dettmers, T., Minervini, P., Stenetorp, P., and Riedel, S.
\newblock Convolutional 2d knowledge graph embeddings.
\newblock In \emph{Proceedings of the Thirty-Second {AAAI} Conference on Artificial Intelligence, (AAAI-18), New Orleans, Louisiana, USA, February 2-7, 2018}, pp.\  1811--1818, 2018.

\bibitem[Devlin et~al.(2018)Devlin, Chang, Lee, and Toutanova]{Bert}
Devlin, J., Chang, M., Lee, K., and Toutanova, K.
\newblock {BERT:} pre-training of deep bidirectional transformers for language understanding.
\newblock \emph{CoRR}, abs/1810.04805, 2018.
\newblock URL \url{http://arxiv.org/abs/1810.04805}.

\bibitem[Dobson \& Doig(2003)Dobson and Doig]{dobson2003distinguishing}
Dobson, P.~D. and Doig, A.~J.
\newblock Distinguishing enzyme structures from non-enzymes without alignments.
\newblock \emph{Journal of molecular biology}, 330\penalty0 (4):\penalty0 771--783, 2003.

\bibitem[Fang et~al.(2023)Fang, Zhang, YANG, Wang, and Chen]{gpfplus2023}
Fang, T., Zhang, Y., YANG, Y., Wang, C., and Chen, L.
\newblock Universal prompt tuning for graph neural networks.
\newblock In Oh, A., Naumann, T., Globerson, A., Saenko, K., Hardt, M., and Levine, S. (eds.), \emph{Advances in Neural Information Processing Systems}, volume~36, pp.\  52464--52489. Curran Associates, Inc., 2023.

\bibitem[Galkin et~al.(2022)Galkin, Denis, Wu, and Hamilton]{NodePiece}
Galkin, M., Denis, E.~G., Wu, J., and Hamilton, W.~L.
\newblock Nodepiece: Compositional and parameter-efficient representations of large knowledge graphs.
\newblock In \emph{The Tenth International Conference on Learning Representations, {ICLR} 2022, Virtual Event, April 25-29, 2022}. OpenReview.net, 2022.

\bibitem[Galkin et~al.(2023)Galkin, Yuan, Mostafa, Tang, and Zhu]{ultra}
Galkin, M., Yuan, X., Mostafa, H., Tang, J., and Zhu, Z.
\newblock Towards foundation models for knowledge graph reasoning.
\newblock \emph{CoRR}, abs/2310.04562, 2023.
\newblock URL \url{https://doi.org/10.48550/arXiv.2310.04562}.

\bibitem[Gao et~al.(2023)Gao, Zhou, and Ribeiro]{isdea}
Gao, J., Zhou, Y., and Ribeiro, B.
\newblock Double permutation equivariance for knowledge graph completion.
\newblock \emph{arXiv preprint arXiv:2302.01313}, 2023.
\newblock URL \url{https://doi.org/10.48550/arXiv.2302.01313}.

\bibitem[Ge et~al.(2023)Ge, Zhao, Liu, Cheng, Li, Wang, and Yin]{ge2023enhancing}
Ge, Q., Zhao, Z., Liu, Y., Cheng, A., Li, X., Wang, S., and Yin, D.
\newblock Enhancing graph neural networks with structure-based prompt.
\newblock \emph{CoRR}, abs/2310.17394, 2023.

\bibitem[Geng et~al.(2023)Geng, Chen, Pan, Chen, Jiang, Zhang, and Chen]{rmpi}
Geng, Y., Chen, J., Pan, J.~Z., Chen, M., Jiang, S., Zhang, W., and Chen, H.
\newblock Relational message passing for fully inductive knowledge graph completion.
\newblock In \emph{2023 IEEE 39th International Conference on Data Engineering (ICDE)}, pp.\  1221--1233. IEEE, 2023.
\newblock URL \url{https://doi.org/10.1109/ICDE55515.2023.00098}.

\bibitem[Gesese et~al.(2022)Gesese, Sack, and Alam]{TextKGR-RAILD-IJCKG22}
Gesese, G.~A., Sack, H., and Alam, M.
\newblock {RAILD:} towards leveraging relation features for inductive link prediction in knowledge graphs.
\newblock In Artale, A., Calvanese, D., Wang, H., and Zhang, X. (eds.), \emph{Proceedings of the 11th International Joint Conference on Knowledge Graphs, {IJCKG} 2022, Hangzhou, China, October 27-28, 2022}, pp.\  82--90. {ACM}, 2022.
\newblock URL \url{https://doi.org/10.1145/3579051.3579066}.

\bibitem[Gong et~al.(2023)Gong, Li, Yu, Yao, Tan, Yu, and Yin]{gong2023prompt}
Gong, C., Li, X., Yu, J., Yao, C., Tan, J., Yu, C., and Yin, D.
\newblock Prompt {Tuning} for {Multi}-{View} {Graph} {Contrastive} {Learning}.
\newblock \emph{arXiv preprint}, 2023.

\bibitem[Hamilton et~al.(2017)Hamilton, Ying, and Leskovec]{GraphSAGE}
Hamilton, W.~L., Ying, Z., and Leskovec, J.
\newblock Inductive representation learning on large graphs.
\newblock In \emph{Advances in Neural Information Processing Systems 30: Annual Conference on Neural Information Processing Systems 2017, December 4-9, 2017, Long Beach, CA, {USA}}, pp.\  1024--1034, 2017.

\bibitem[Huang et~al.(2023{\natexlab{a}})Huang, Ren, Chen, Kržmanc, Zeng, Liang, and Leskovec]{huang2023prodigy}
Huang, Q., Ren, H., Chen, P., Kržmanc, G., Zeng, D., Liang, P., and Leskovec, J.
\newblock {PRODIGY}: {Enabling} {In}-context {Learning} {Over} {Graphs}.
\newblock In \emph{{NeurIPS}}, 2023{\natexlab{a}}.

\bibitem[Huang et~al.(2023{\natexlab{b}})Huang, Romero, Ceylan, and Barcel{\'{o}}]{LPTheory-NIPS23}
Huang, X., Romero, M., Ceylan, {\.I}.~{\.I}., and Barcel{\'{o}}, P.
\newblock A theory of link prediction via relational weisfeiler-leman on knowledge graphs.
\newblock In \emph{Advances in Neural Information Processing Systems 36: Annual Conference on Neural Information Processing Systems 2023, NeurIPS 2023, New Orleans, LA, USA, December 10 - 16, 2023}, 2023{\natexlab{b}}.

\bibitem[Kipf \& Welling(2017)Kipf and Welling]{GCN-ICLR17}
Kipf, T.~N. and Welling, M.
\newblock Semi-supervised classification with graph convolutional networks.
\newblock In \emph{5th International Conference on Learning Representations, {ICLR} 2017, Toulon, France, April 24-26, 2017, Conference Track Proceedings}. OpenReview.net, 2017.

\bibitem[Kriege \& Mutzel(2012)Kriege and Mutzel]{MUTAG}
Kriege, N. and Mutzel, P.
\newblock Subgraph matching kernels for attributed graphs.
\newblock In \emph{Proceedings of the 29th International Coference on International Conference on Machine Learning}, ICML'12, pp.\  291–298, Madison, WI, USA, 2012. Omnipress.
\newblock ISBN 9781450312851.

\bibitem[Lee et~al.(2023)Lee, Chung, and Whang]{ingram}
Lee, J., Chung, C., and Whang, J.~J.
\newblock {I}n{G}ram: Inductive knowledge graph embedding via relation graphs.
\newblock In \emph{Proceedings of the 40th International Conference on Machine Learning}, volume 202, pp.\  18796--18809. PMLR, 23--29 Jul 2023.
\newblock URL \url{https://proceedings.mlr.press/v202/lee23c.html}.

\bibitem[Li et~al.(2024)Li, Wang, Li, Yu, and Li]{zerog}
Li, Y., Wang, P., Li, Z., Yu, J.~X., and Li, J.
\newblock Zerog: Investigating cross-dataset zero-shot transferability in graphs.
\newblock In \emph{Proceedings of the 30th ACM SIGKDD Conference on Knowledge Discovery and Data Mining}, pp.\  1725--1735, 2024.

\bibitem[Liu et~al.(2023{\natexlab{a}})Liu, Peng, Xu, and Peng]{IKR-JWWW23}
Liu, B., Peng, M., Xu, W., and Peng, M.
\newblock Neighboring relation enhanced inductive knowledge graph link prediction via meta-learning.
\newblock \emph{World Wide Web {(WWW)}}, 26\penalty0 (5):\penalty0 2909--2930, 2023{\natexlab{a}}.

\bibitem[Liu et~al.(2024)Liu, Feng, Kong, Liang, Tao, Chen, and Zhang]{OFA}
Liu, H., Feng, J., Kong, L., Liang, N., Tao, D., Chen, Y., and Zhang, M.
\newblock One for all: Towards training one graph model for all classification tasks.
\newblock \emph{ICLR}, 2024.

\bibitem[Liu et~al.(2023{\natexlab{b}})Liu, Yang, Lu, Chen, Li, Zhang, Bai, Fang, Sun, Yu, and Shi]{GFMSurvey-arxiv23}
Liu, J., Yang, C., Lu, Z., Chen, J., Li, Y., Zhang, M., Bai, T., Fang, Y., Sun, L., Yu, P.~S., and Shi, C.
\newblock Towards graph foundation models: {A} survey and beyond.
\newblock \emph{CoRR}, abs/2310.11829, 2023{\natexlab{b}}.
\newblock \doi {10.48550/arXiv.2310.11829}.

\bibitem[Liu et~al.(2023{\natexlab{c}})Liu, Yu, Fang, and Zhang]{liu2023graphprompt}
Liu, Z., Yu, X., Fang, Y., and Zhang, X.
\newblock Graphprompt: {Unifying} {Pre}-{Training} and {Downstream} {Tasks} for {Graph} {Neural} {Networks}.
\newblock In \emph{The {Web} {Conference}}, pp.\  417--428, 2023{\natexlab{c}}.

\bibitem[Mao et~al.(2024)Mao, Chen, Tang, Zhao, Ma, Zhao, Shah, Galkin, and Tang]{GFMSurvey-arxiv24}
Mao, H., Chen, Z., Tang, W., Zhao, J., Ma, Y., Zhao, T., Shah, N., Galkin, M., and Tang, J.
\newblock Graph foundation models.
\newblock \emph{CoRR}, abs/2402.02216, 2024.
\newblock \doi {10.48550/arXiv.2402.02216}.

\bibitem[Mikolov et~al.(2013)Mikolov, Chen, Corrado, and Dean]{word2vec}
Mikolov, T., Chen, K., Corrado, G., and Dean, J.
\newblock Efficient estimation of word representations in vector space.
\newblock In \emph{Proceedings of the 1st International Conference on Learning Representations, {ICLR} 2013, Scottsdale, Arizona, USA, May 2-4, 2013, Workshop Track Proceedings}, 2013.

\bibitem[Namata et~al.(2012)Namata, London, Getoor, Huang, and Edu]{namata2012homoPubmed}
Namata, G., London, B., Getoor, L., Huang, B., and Edu, U.
\newblock Query-driven active surveying for collective classification.
\newblock In \emph{10th international workshop on mining and learning with graphs}, volume~8, pp.\ ~1, 2012.

\bibitem[Pei et~al.(2019)Pei, Wei, Chang, Lei, and Yang]{pei2019heterophilicDatasets}
Pei, H., Wei, B., Chang, K. C.-C., Lei, Y., and Yang, B.
\newblock Geom-gcn: Geometric graph convolutional networks.
\newblock In \emph{International Conference on Learning Representations}, 2019.

\bibitem[Rosenstein et~al.(2005)Rosenstein, Marx, Kaelbling, and Dietterich]{rosenstein2005transfer}
Rosenstein, M.~T., Marx, Z., Kaelbling, L.~P., and Dietterich, T.~G.
\newblock To transfer or not to transfer.
\newblock In \emph{{NeurIPS}}, volume 898, 2005.

\bibitem[Rossi \& Ahmed(2015)Rossi and Ahmed]{rossi2015network}
Rossi, R. and Ahmed, N.
\newblock The network data repository with interactive graph analytics and visualization.
\newblock In \emph{Proceedings of the AAAI conference on artificial intelligence}, volume~29, 2015.

\bibitem[Sadeghian et~al.(2019)Sadeghian, Armandpour, Ding, and Wang]{DRUM-NIPS19}
Sadeghian, A., Armandpour, M., Ding, P., and Wang, D.~Z.
\newblock {DRUM:} end-to-end differentiable rule mining on knowledge graphs.
\newblock In \emph{Advances in Neural Information Processing Systems 32: Annual Conference on Neural Information Processing Systems 2019, NeurIPS 2019, December 8-14, 2019, Vancouver, BC, Canada}, pp.\  15321--15331, 2019.

\bibitem[Safavi \& Koutra(2020)Safavi and Koutra]{Codex}
Safavi, T. and Koutra, D.
\newblock Codex: {A} comprehensive knowledge graph completion benchmark.
\newblock In \emph{Proceedings of the 2020 Conference on Empirical Methods in Natural Language Processing, {EMNLP} 2020, Online, November 16-20, 2020}, pp.\  8328--8350, 2020.

\bibitem[Sanh et~al.(2019)Sanh, Debut, Chaumond, and Wolf]{DistilBert}
Sanh, V., Debut, L., Chaumond, J., and Wolf, T.
\newblock Distilbert, a distilled version of {BERT:} smaller, faster, cheaper and lighter.
\newblock \emph{CoRR}, abs/1910.01108, 2019.
\newblock URL \url{http://arxiv.org/abs/1910.01108}.

\bibitem[Schlichtkrull et~al.(2018)Schlichtkrull, Kipf, Bloem, van~den Berg, Titov, and Welling]{RGCN-ESWC18}
Schlichtkrull, M.~S., Kipf, T.~N., Bloem, P., van~den Berg, R., Titov, I., and Welling, M.
\newblock Modeling relational data with graph convolutional networks.
\newblock In \emph{The Semantic Web - 15th International Conference, {ESWC} 2018, Heraklion, Crete, Greece, June 3-7, 2018, Proceedings}, volume 10843 of \emph{Lecture Notes in Computer Science}, pp.\  593--607. Springer, 2018.

\bibitem[Sen et~al.(2008)Sen, Namata, Bilgic, Getoor, Galligher, and Eliassi-Rad]{sen2008homoCoraCiteseer}
Sen, P., Namata, G., Bilgic, M., Getoor, L., Galligher, B., and Eliassi-Rad, T.
\newblock Collective classification in network data.
\newblock \emph{AI magazine}, 29\penalty0 (3):\penalty0 93--93, 2008.

\bibitem[Song et~al.(2020)Song, Tan, Qin, Lu, and Liu]{MPNet}
Song, K., Tan, X., Qin, T., Lu, J., and Liu, T.
\newblock Mpnet: Masked and permuted pre-training for language understanding.
\newblock In Larochelle, H., Ranzato, M., Hadsell, R., Balcan, M., and Lin, H. (eds.), \emph{Advances in Neural Information Processing Systems 33: Annual Conference on Neural Information Processing Systems 2020, NeurIPS 2020, December 6-12, 2020, virtual}, 2020.

\bibitem[Sun et~al.(2022)Sun, Zhou, He, Wang, and Wang]{sun2022gppt}
Sun, M., Zhou, K., He, X., Wang, Y., and Wang, X.
\newblock {GPPT}: {Graph} {Pre}-{Training} and {Prompt} {Tuning} to {Generalize} {Graph} {Neural} {Networks}.
\newblock In \emph{{KDD}}, pp.\  1717--1727, 2022.

\bibitem[Sun et~al.(2023)Sun, Cheng, Li, Liu, and Guan]{sun2023all}
Sun, X., Cheng, H., Li, J., Liu, B., and Guan, J.
\newblock All in one: Multi-task prompting for graph neural networks.
\newblock In \emph{Proceedings of the 29th ACM SIGKDD Conference on Knowledge Discovery and Data Mining}, pp.\  2120--2131, 2023.

\bibitem[Sun et~al.(2019)Sun, Deng, Nie, and Tang]{RotatE}
Sun, Z., Deng, Z., Nie, J., and Tang, J.
\newblock Rotate: Knowledge graph embedding by relational rotation in complex space.
\newblock In \emph{Proceedings of the 7th International Conference on Learning Representations, {ICLR} 2019, New Orleans, LA, USA, May 6-9, 2019}, 2019.

\bibitem[Tan et~al.(2023)Tan, Guo, Ding, and Liu]{tan2023virtual}
Tan, Z., Guo, R., Ding, K., and Liu, H.
\newblock Virtual {Node} {Tuning} for {Few}-shot {Node} {Classification}.
\newblock In \emph{{KDD}}, pp.\  2177--2188, 2023.

\bibitem[Tang et~al.(2009)Tang, Sun, Wang, and Yang]{tang2009social}
Tang, J., Sun, J., Wang, C., and Yang, Z.
\newblock Social influence analysis in large-scale networks.
\newblock In \emph{Proceedings of the 15th ACM SIGKDD international conference on Knowledge discovery and data mining}, pp.\  807--816, 2009.

\bibitem[Tang et~al.(2024)Tang, Yang, Wei, Shi, Xia, Yin, and Huang]{HiGPT}
Tang, J., Yang, Y., Wei, W., Shi, L., Xia, L., Yin, D., and Huang, C.
\newblock Higpt: Heterogeneous graph language model, 2024.
\newblock URL \url{https://arxiv.org/abs/2402.16024}.

\bibitem[Teru et~al.(2020)Teru, Denis, and Hamilton]{GraIL-ICML19}
Teru, K.~K., Denis, E.~G., and Hamilton, W.~L.
\newblock Inductive relation prediction by subgraph reasoning.
\newblock In \emph{Proceedings of the 37th International Conference on Machine Learning, {ICML} 2020, 13-18 July 2020, Virtual Event}, volume 119 of \emph{Proceedings of Machine Learning Research}, pp.\  9448--9457. {PMLR}, 2020.

\bibitem[Toutanova \& Chen(2015)Toutanova and Chen]{FB15k237}
Toutanova, K. and Chen, D.
\newblock Observed versus latent features for knowledge base and text inference.
\newblock In \emph{Proceedings of the 3rd Workshop on Continuous Vector Space Models and their Compositionality}, pp.\  57--66, 2015.

\bibitem[Vashishth et~al.(2020)Vashishth, Sanyal, Nitin, and Talukdar]{CompGCN-ICLR20}
Vashishth, S., Sanyal, S., Nitin, V., and Talukdar, P.~P.
\newblock Composition-based multi-relational graph convolutional networks.
\newblock In \emph{8th International Conference on Learning Representations, {ICLR} 2020, Addis Ababa, Ethiopia, April 26-30, 2020}. OpenReview.net, 2020.

\bibitem[Velickovic et~al.(2018)Velickovic, Cucurull, Casanova, Romero, Li{\`{o}}, and Bengio]{GAT-ICLR18}
Velickovic, P., Cucurull, G., Casanova, A., Romero, A., Li{\`{o}}, P., and Bengio, Y.
\newblock Graph attention networks.
\newblock 2018.

\bibitem[Wang et~al.(2021{\natexlab{a}})Wang, Wang, Huang, You, Leskovec, and Kuo]{InductivE-IJCNN21}
Wang, B., Wang, G., Huang, J., You, J., Leskovec, J., and Kuo, C.~J.
\newblock Inductive learning on commonsense knowledge graph completion.
\newblock In \emph{International Joint Conference on Neural Networks, {IJCNN} 2021, Shenzhen, China, July 18-22, 2021}, pp.\  1--8. {IEEE}, 2021{\natexlab{a}}.
\newblock URL \url{https://doi.org/10.1109/IJCNN52387.2021.9534355}.

\bibitem[Wang et~al.(2021{\natexlab{b}})Wang, Liu, Ma, and Sheng]{OurMulDE}
Wang, K., Liu, Y., Ma, Q., and Sheng, Q.~Z.
\newblock Mulde: Multi-teacher knowledge distillation for low-dimensional knowledge graph embeddings.
\newblock In \emph{Proceedings of the {WWW} '21: The Web Conference 2021, Virtual Event / Ljubljana, Slovenia, April 19-23, 2021}, pp.\  1716--1726, 2021{\natexlab{b}}.

\bibitem[Wang et~al.(2022{\natexlab{a}})Wang, Liu, and Sheng]{OurHaLE}
Wang, K., Liu, Y., and Sheng, Q.~Z.
\newblock Swift and sure: Hardness-aware contrastive learning for low-dimensional knowledge graph embeddings.
\newblock In \emph{{WWW} '22: The {ACM} Web Conference 2022, Virtual Event, Lyon, France, April 25 - 29, 2022}, pp.\  838--849. {ACM}, 2022{\natexlab{a}}.

\bibitem[Wang et~al.(2024{\natexlab{a}})Wang, Xu, and Luo]{OurTIGER}
Wang, K., Xu, Y., and Luo, S.
\newblock {TIGER:} training inductive graph neural network for large-scale knowledge graph reasoning.
\newblock \emph{Proc. {VLDB} Endow.}, 17\penalty0 (10):\penalty0 2459--2472, 2024{\natexlab{a}}.

\bibitem[Wang et~al.(2024{\natexlab{b}})Wang, Xu, Wu, and Luo]{OurProLINK}
Wang, K., Xu, Y., Wu, Z., and Luo, S.
\newblock {LLM} as prompter: Low-resource inductive reasoning on arbitrary knowledge graphs.
\newblock In Ku, L.-W., Martins, A., and Srikumar, V. (eds.), \emph{Findings of the Association for Computational Linguistics ACL 2024}, pp.\  3742--3759, Bangkok, Thailand and virtual meeting, August 2024{\natexlab{b}}. Association for Computational Linguistics.
\newblock URL \url{https://aclanthology.org/2024.findings-acl.224}.

\bibitem[Wang et~al.(2025{\natexlab{a}})Wang, Lin, and Luo]{OurGraPE}
Wang, K., Lin, D., and Luo, S.
\newblock Graph percolation embeddings for efficient knowledge graph inductive reasoning.
\newblock \emph{IEEE Transactions on Knowledge and Data Engineering}, 37\penalty0 (3):\penalty0 1198--1212, 2025{\natexlab{a}}.
\newblock \doi{10.1109/TKDE.2024.3508064}.

\bibitem[Wang et~al.(2017)Wang, Mao, Wang, and Guo]{2017Survey}
Wang, Q., Mao, Z., Wang, B., and Guo, L.
\newblock Knowledge graph embedding: A survey of approaches and applications.
\newblock \emph{IEEE Transactions on Knowledge and Data Engineering}, 29\penalty0 (12):\penalty0 2724--2743, 2017.

\bibitem[Wang et~al.(2022{\natexlab{b}})Wang, Dong, Huang, Chen, and Li]{wang2022faith}
Wang, S., Dong, Y., Huang, X., Chen, C., and Li, J.
\newblock Faith: Few-shot graph classification with hierarchical task graphs.
\newblock \emph{arXiv preprint arXiv:2205.02435}, 2022{\natexlab{b}}.

\bibitem[Wang et~al.(2020)Wang, Wei, Dong, Bao, Yang, and Zhou]{MiniLM}
Wang, W., Wei, F., Dong, L., Bao, H., Yang, N., and Zhou, M.
\newblock Minilm: Deep self-attention distillation for task-agnostic compression of pre-trained transformers.
\newblock In Larochelle, H., Ranzato, M., Hadsell, R., Balcan, M., and Lin, H. (eds.), \emph{Advances in Neural Information Processing Systems 33: Annual Conference on Neural Information Processing Systems 2020, NeurIPS 2020, December 6-12, 2020, virtual}, 2020.

\bibitem[Wang et~al.(2025{\natexlab{b}})Wang, Fan, Wang, and Ma]{GFM_HKPU_arxiv2503}
Wang, Y., Fan, W., Wang, S., and Ma, Y.
\newblock Towards graph foundation models: {A} transferability perspective.
\newblock \emph{CoRR}, abs/2503.09363, 2025{\natexlab{b}}.
\newblock \doi{10.48550/ARXIV.2503.09363}.
\newblock URL \url{https://doi.org/10.48550/arXiv.2503.09363}.

\bibitem[Wu et~al.(2019)Wu, Souza, Zhang, Fifty, Yu, and Weinberger]{wu2019simplifying}
Wu, F., Souza, A., Zhang, T., Fifty, C., Yu, T., and Weinberger, K.
\newblock Simplifying graph convolutional networks.
\newblock In \emph{International conference on machine learning}, pp.\  6861--6871. PMLR, 2019.

\bibitem[Xia \& Huang(2024)Xia and Huang]{AnyGraph}
Xia, L. and Huang, C.
\newblock Anygraph: Graph foundation model in the wild, 2024.
\newblock URL \url{https://arxiv.org/abs/2408.10700}.

\bibitem[Xia et~al.(2024)Xia, Kao, and Huang]{OpenGraph}
Xia, L., Kao, B., and Huang, C.
\newblock Opengraph: Towards open graph foundation models, 2024.
\newblock URL \url{https://arxiv.org/abs/2403.01121}.

\bibitem[Xiong et~al.(2017)Xiong, Hoang, and Wang]{NELL995}
Xiong, W., Hoang, T., and Wang, W.~Y.
\newblock Deeppath: {A} reinforcement learning method for knowledge graph reasoning.
\newblock In \emph{Proceedings of the 2017 Conference on Empirical Methods in Natural Language Processing, {EMNLP} 2017, Copenhagen, Denmark, September 9-11, 2017}, pp.\  564--573. Association for Computational Linguistics, 2017.

\bibitem[Yan et~al.(2022)Yan, Ma, Gao, Tang, and Chen]{Cycle-ICML22}
Yan, Z., Ma, T., Gao, L., Tang, Z., and Chen, C.
\newblock Cycle representation learning for inductive relation prediction.
\newblock In \emph{International Conference on Machine Learning, {ICML} 2022, 17-23 July 2022, Baltimore, Maryland, {USA}}, volume 162 of \emph{Proceedings of Machine Learning Research}, pp.\  24895--24910. {PMLR}, 2022.
\newblock URL \url{https://proceedings.mlr.press/v162/yan22a.html}.

\bibitem[Yanardag \& Vishwanathan(2015)Yanardag and Vishwanathan]{yanardag2015deep}
Yanardag, P. and Vishwanathan, S.
\newblock Deep graph kernels.
\newblock In \emph{Proceedings of the 21th ACM SIGKDD international conference on knowledge discovery and data mining}, pp.\  1365--1374, 2015.

\bibitem[Yang et~al.(2015)Yang, Yih, He, Gao, and Deng]{DistMult}
Yang, B., Yih, W., He, X., Gao, J., and Deng, L.
\newblock Embedding entities and relations for learning and inference in knowledge bases.
\newblock In \emph{Proceedings of the 3rd International Conference on Learning Representations, {ICLR} 2015, San Diego, CA, USA, May 7-9, 2015}, 2015.

\bibitem[Ye et~al.(2024)Ye, Zhang, Wang, Xu, and Zhang]{InstructGLM}
Ye, R., Zhang, C., Wang, R., Xu, S., and Zhang, Y.
\newblock Language is all a graph needs, 2024.
\newblock URL \url{https://arxiv.org/abs/2308.07134}.

\bibitem[Zhang et~al.(2020)Zhang, Li, Xia, Wang, and Jin]{RevisitGNN}
Zhang, M., Li, P., Xia, Y., Wang, K., and Jin, L.
\newblock Labeling trick: A theory of using graph neural networks for multi-node representation learning.
\newblock \emph{CoRR}, abs/2010.16103, 2020.
\newblock URL \url{https://arxiv.org/abs/2010.16103}.

\bibitem[Zhang \& Yao(2022)Zhang and Yao]{REDGNN-WWW22}
Zhang, Y. and Yao, Q.
\newblock Knowledge graph reasoning with relational digraph.
\newblock In \emph{{WWW} '22: The {ACM} Web Conference 2022, Virtual Event, Lyon, France, April 25 - 29, 2022}, pp.\  912--924. {ACM}, 2022.

\bibitem[Zhang et~al.(2023)Zhang, Zhou, Yao, Chu, and Han]{Adaprop}
Zhang, Y., Zhou, Z., Yao, Q., Chu, X., and Han, B.
\newblock Adaprop: Learning adaptive propagation for graph neural network based knowledge graph reasoning.
\newblock In \emph{Proceedings of the 29th {ACM} {SIGKDD} Conference on Knowledge Discovery and Data Mining, {KDD} 2023, Long Beach, CA, USA, August 6-10, 2023}, pp.\  3446--3457. {ACM}, 2023.
\newblock \doi{10.1145/3580305.3599404}.
\newblock URL \url{https://doi.org/10.1145/3580305.3599404}.

\bibitem[Zhang et~al.(2025)Zhang, Bevilacqua, Galkin, and Ribeiro]{TRIX}
Zhang, Y., Bevilacqua, B., Galkin, M., and Ribeiro, B.
\newblock {TRIX:} {A} more expressive model for zero-shot domain transfer in knowledge graphs.
\newblock \emph{CoRR}, abs/2502.19512, 2025.
\newblock \doi{10.48550/ARXIV.2502.19512}.
\newblock URL \url{https://doi.org/10.48550/arXiv.2502.19512}.

\bibitem[Zhao et~al.(2024)Zhao, Mostafa, Galkin, Bronstein, Zhu, and Tang]{GraphAny-Arxiv24}
Zhao, J., Mostafa, H., Galkin, M., Bronstein, M., Zhu, Z., and Tang, J.
\newblock Graphany: A foundation model for node classification on any graph, 2024.
\newblock URL \url{https://arxiv.org/abs/2405.20445}.

\bibitem[Zhou et~al.(2023)Zhou, Bevilacqua, and Ribeiro]{mtdea}
Zhou, J., Bevilacqua, B., and Ribeiro, B.
\newblock An ood multi-task perspective for link prediction with new relation types and nodes.
\newblock \emph{arXiv preprint arXiv:2307.06046}, 2023.
\newblock URL \url{https://doi.org/10.48550/arXiv.2307.06046}.

\bibitem[Zhu et~al.(2021)Zhu, Zhang, Xhonneux, and Tang]{NBF-NIPS21}
Zhu, Z., Zhang, Z., Xhonneux, L. A.~C., and Tang, J.
\newblock Neural bellman-ford networks: {A} general graph neural network framework for link prediction.
\newblock In \emph{Advances in Neural Information Processing Systems 34: Annual Conference on Neural Information Processing Systems 2021, NeurIPS 2021, December 6-14, 2021, virtual}, pp.\  29476--29490, 2021.

\bibitem[Zhu et~al.(2023)Zhu, Yuan, Galkin, Xhonneux, Zhang, Gazeau, and Tang]{ANet}
Zhu, Z., Yuan, X., Galkin, M., Xhonneux, L. A.~C., Zhang, M., Gazeau, M., and Tang, J.
\newblock A*net: {A} scalable path-based reasoning approach for knowledge graphs.
\newblock In \emph{Advances in Neural Information Processing Systems 36: Annual Conference on Neural Information Processing Systems 2023, NeurIPS 2023, New Orleans, LA, USA, December 10 - 16, 2023}, 2023.

\bibitem[Zi et~al.(2024)Zi, Zhao, Sun, Lin, Cheng, and Li]{ProG}
Zi, C., Zhao, H., Sun, X., Lin, Y., Cheng, H., and Li, J.
\newblock Prog: A graph prompt learning benchmark.
\newblock \emph{arXiv preprint arXiv:2406.05346}, 2024.

\end{thebibliography}


\clearpage
\appendix
\section*{Appendix}

\section{Notations and Definitions} 
\label{app_sec:notations}
The notations used in this paper and their descriptions are summarized in Table~\ref{tab:notations}.

\begin{table}[h]
\caption{Summary of the major notations in this paper.}
\centering
\small
\label{tab:notations}
\begin{tabular}{l|l}
\hline
\textbf{Symbol} & \textbf{Description} \\[3pt]
\hline
$\mathcal{G}$ & A knowledge graph (KG) \\[3pt]
$\mathcal{T}$ & The set of triples in a KG\\[3pt]
$\mathcal{E},\mathcal{R}$ & The entity set and relation set in a KG\\[3pt]
$|\mathcal{T}|, |\mathcal{E}|, |\mathcal{R}|$ & The item number in a specific set\\ [3pt]
$e,r$ & An entity (e) or a relation (r) in a KG\\[3pt]
$\bm{q}=(e_q,r_q)$ & A query with an entity $e_q$ and a relation $r_q$ \\[3pt]
$e_a$ & The ground-truth entity of a query \\[3pt]
$\features$ & node/entity feature matrix \\[3pt]
$d^0, d$ & Dimension of features and embeddings \\[3pt]
$\gtrain, \ginf$ & The training KG and inference KG \\[3pt]
\hline
${\it CMP}()$ & A conditional message passing module \\[3pt] 
$\vh_{v}$ & The representation of $e_v$ conditioned on $\bm{q}$ \\[3pt]
$\mathbf{P}$ & Relation-level interaction representations\\[3pt]
$\mathbf{R}_{g}$ & Query-independent relation representations\\[3pt]
$\mathbf{R}_{q}$ & Relation representations conditioned on $\bm{q}$ \\[3pt]
$\mathbf{H}_{g}$ & Query-independent entity representations\\[3pt]
$\mathbf{H}_{q}$ & Entity representations conditioned on $\bm{q}$ \\[3pt]
$\init(\bm{q}, e_v)$ & The initialization function of CMP  \\[3pt]
$\mes()$ & The message passing function of CMP \\[3pt]
$\aggregate(), \updatefunc()$ & The aggretation and update functions \\[3pt]
$\mathcal{N}_r(e_v)$ & The direct neighbors of $e_v$ connecting via $r$ \\[3pt]
\hline
$p(\bm{q}, e_v)$ & The plausibility score for the triple \\[3pt] 
$\tilde{\features}$ & The unified feature matrix \\[3pt]
$\graph_r$ & The relation graph corresponding to $\graph$ \\[3pt]
$\mathcal{R}_{\it fund}$ & The interaction types in the relation graph \\[3pt]
$\triples_{\it r}$ & The triples in the relation graph \\[3pt]
$\mathcal{S}_{e_v}$ & The semantic neighbors of $e_v$ \\[3pt]
$\mathbbm{1}(q=v)$ & The indicator function \\[3pt] 
$\mathcal{L}$ & The loss function of \method \\[3pt]
$\mathcal{F}$ & The set of diverse features for training \\[3pt]
\hline
\end{tabular}
\vspace{-3mm}
\end{table}

\section{Tasks and Datasets}
\label{app:data}

\subsection{Task-Specific Datasets}

\textbf{Link-level KG Reasoning:} 
We conduct inductive KG reasoning experiments on 24 datasets.
Half of them are derived from the GraIL work~\citep{GraIL-ICML19}, which are constructed from commonly-used KG benchmarks, including WN18RR~\citep{WN18RR}, FB15k237 \citep{FB15k237} and NELL-995 \citep{NELL995}. 
In these datasets, the train graphs and the test graphs share the same relation types.
To evaluate performance in the full context of inductive reasoning,
we also employ 12 datasets used in the InGram work ~\citep{ingram}.
The InGram datasets were derived from three real-world KG benchmarks: FB15k237 \citep{FB15k237}, Wikidata68K~\citep{TextKGR-RAILD-IJCKG22}, and NELL-995~\citep{NELL995}. 
There are four datasets in each series with different proportions of triplets with new relations as 100\%, 75\%, 50\%, and 25\%.
Note that, there are other KG datasets where textual descriptions are not easily accessible; we leave the evaluation of such datasets for future work.
Structural statistics for these datasets can be found in Table \ref{table:ds-test}.

\noindent\textbf{Node/Graph-level Classification:} 
To evaluate the adaptability of our method across various types/domains of graph tasks, we conduct experiments on 14 datasets involving both node-level and graph-level classification tasks. Following prior studies~\citep{ProG}, we employ six commonly-used node classification datasets, including homophilic graph datasets (Cora, Citeseer, PubMed)~\citep{sen2008homoCoraCiteseer,namata2012homoPubmed}, and heterophilic graph datasets (Wisconsin, Texas, Actor)~\citep{pei2019heterophilicDatasets,tang2009social}. 
Furthermore, we considered eight graph classification datasets from various domains, such as social networks (IMDB-B, COLLAB)~\citep{yanardag2015deep}, biological datasets (ENZYMES, PROTEINS, DD)~\citep{dobson2003distinguishing,borgwardt2005protein,wang2022faith}, and small molecule datasets (MUTAG, COX2, BZR)~\citep{MUTAG, rossi2015network}. More detailed information on these datasets can be found in Table~\ref{table:ds-class}.

\begin{table}[!ht]
\small
\centering
\caption{Statistics of pre-training KG datasets.}
\label{table:ds-train}
\setlength\tabcolsep{2pt}
\begin{tabular}{lccccc}
\hline
\multirow{2}{*}{Dataset} & \multirow{2}{*}{$|\etrain|$} & \multirow{2}{*}{$|\rtrain|$} & \multicolumn{3}{c}{$|\ttrain|$} \\
& & & {\#Train} & {\#Validation} & {\#Test} \\
\hline
WN18RR & 40.9k & 11 & 86.8k & 3.0k & 3.1k \\
FB15k-237 & 14.5k & 237 & 272.1k & 17.5k & 20.4k \\
CodexMedium & 17.0k & 51 & 185.5k & 10.3k & 10.3k \\
\hline
\end{tabular}
\end{table}

\begin{table}[!ht]
\centering
\small
\caption{Statistics of Bert-based text encoder.}
\label{table:model-bert}
\begin{tabular}{lll}
\hline
Method & Dim & Model Name \\
\hline
Bert & 768 & bert-base-nli-mean-tokens \\
MPNet & 768 & all-mpnet-base-v2 \\
MiniLM & 384 & paraphrase-MiniLM-L6-v2 \\
DistilBert & 512 & distiluse-base-multilingual-cased-v1 \\
\hline
\end{tabular}
\end{table}

\subsection{Task-Specific Data Preparation}

To evaluate the generalizability of \method as a foundational graph reasoning engine, we reformulate various task data into a unified KG reasoning format, which includes a multi-relational graph structure and semantic node features. We illustrate the task-specific data forms in Figure \ref{fig:3}.

\noindent\textbf{Link-level KG Reasoning:}
For graph structure, KG triples can be directly represented as a multi-relational graph structure. For semantic features, we generate entity-level embeddings from available textual descriptions using BERT-based sentence encoders. To simulate different feature spaces, we employ four classical language models as sentence encoders, including BERT~\citep{Bert}, MPNet~\citep{MPNet}, MiniLM~\citep{MiniLM}, and DistilBERT~\citep{DistilBert}. MiniLM and DistilBERT have different embedding dimensions compared to BERT and MPNet. Details of textual encoders are provided in Table~\ref{table:model-bert} in the Appendix.

\noindent\textbf{Node-level Classification:}
For semantic features, we directly utilize the provided input node features. In terms of graph structure, we augment the original homogeneous graph by introducing ``label classes'' as distinct nodes. Edges with a ``labeling'' relation type are added to connect nodes with training labels to their corresponding class nodes. Consequently, the augmented graph contains two relation types and two entity types (nodes and labels). This approach eliminates the need to learn specific parameters for each class, enabling support for new nodes or labels through a zero-shot classification paradigm. 
Prior work~\citep{AnyGraph,OpenGraph} adopted a similar format but in a homogeneous graph, lacking the relation distinguishability between original edges and labeling edges.

\noindent\textbf{Graph-level Classification:} 
Unlike node-level classification, graph classification tasks aim to predict the category of an entire graph. The graphs in the training set have no direct connections to the test graphs, which presents a challenge to the CMP reasoning process. 
To address task requirements, both the graph structure and the semantic feature format are specifically designed. First, we integrate all individual graphs into a single disconnected large graph, adding a ``graph'' node representing each graph, which connects to its corresponding nodes via a new relation type.
Next, we perform global semantic encoding of \method on this large graph to obtain the global representations of each graph node, and then capture semantically similar edges among graph nodes. This allows us to convert the task into a node-level task, where the global representations serve as semantic features and reasoning occurs on an augmented graph encompassing all graph nodes and labels.

\section{Semantic-Augmented Relation Graph}    
\label{app:sarg}

CMP-based models eliminate the necessity for learning unique embeddings for each entity. Instead, they depend on trainable relational representations to facilitate relation-specific message functions. To accommodate varied relational vocabularies in new KGs, recent research~\citep{rmpi,mtdea,isdea} emphasizes the importance of identifying the ``invariance'' present in the KG relational structure, thereby enabling any new relation type to be represented using a predefined set of parameters.

Drawing from insights in prior research~\citep{ultra}, we construct a relation graph $\graph_{\it r} = \{\rels, \mathcal{R}_{\it fund}, \triples_{\it r}\}$, where the nodes represent the relations in $\graph$, and the edges $\mathcal{R}_{\it fund}$ capture four types of interactions between relations: "head-to-head", "tail-to-tail", "head-to-tail", and "tail-to-head". For instance, if two triples $(e_1, r_1, e_2)$ and $(e_2, r_2, e_3)$ are linked tail-to-head, an edge $(r_1, \text{``t-h''}, r_2)$ would be added to $\triples_{\it r}$. Since these four interaction types are inherently derived from the triple structure in knowledge graphs, the pre-trained embeddings of these interaction types can be universally shared across KGs, allowing for the parameterization of any unseen relations.

In our \method framework, we refine the relation graph by supplementing the original triple data $\triples$ with additional edges obtained through semantic augmentation.
Specifically, we derive semantic interactions among entities from the unified features $\tilde{\bm{X}}$. For each entity $e_v$, we identify the top $k$ spatially nearest entities in the unified feature space via pairwise similarities, while excluding its direct topological neighbors. The set of semantic neighbors $\mathcal{S}_{e_v}$ is defined as follows:
\begin{align}
\small
\begin{split}
\mathcal{S}_{e_v} = \{e_i \in \ents \mid e_i \in f_s(\tilde{\features}, e_v, k, \delta) \land \ e_i \notin \mathcal{N}_{e_v} \},
\end{split}
\end{align}
Here, $f_s(\cdot)$ represents the similarity function, $\mathcal{N}_{e_v} = \{e_i \in \ents \mid (e_v, r, e_i) \in \triples, r\in\rels\}$ refers to the topological neighbor set of $e_v$. 
The hyperparameters $k$ and $\delta$ refer to the number of neighbors and the similarity threshold, respectively.
The semantic interaction between $e_v$ and each element in $\mathcal{S}_{e_v}$ is regraded as an additional relation type $r_s$. 
Finally, the construction rules for the relation graph $\graph_{\it r}$ can be formalized as follows:
\begin{align}
\small
\begin{split}
\exists e_v \in \ents, r_1, r_2 \in \rels: \quad\quad\quad\quad\quad\quad\quad\quad\quad\quad\quad\quad\quad\quad & \nonumber 
\end{split}
\\
\small
\begin{split}
\exists \ e \in \tilde{\mathcal{N}}_{r_1}^{l_1} \cap \tilde{\mathcal{N}}_{r_2}^{l_2} 
   \ \Rightarrow \ (r_1, l_1\text{-}l_2, r_2), (r_2, l_2\text{-}l_1, r_1) & \in \triples_{\it r}, \nonumber
\end{split}
\\
\small
\begin{split}
\exists \ e \in \mathcal{S}_{e_v} \cap \tilde{\mathcal{N}}_{r_1}^{l_1} 
   \ \Rightarrow \ (r_1, l_1\text{-}\text{`t'}, r_s), (r_s, \text{`t'}\text{-}l_1, r_1) & \in \triples_{\it r}, \nonumber
\end{split}
\\
\small
\begin{split}
\mathcal{S}_{e_v} \neq \emptyset \land e_v \in \tilde{\mathcal{N}}_{r_1}^{l_1} 
   \ \Rightarrow \ (r_1, l_1\text{-}\text{`h'}, r_s), (r_s, \text{`h'}\text{-}l_1, r_1) & \in \triples_{\it r}, \nonumber
\end{split}
\end{align}
where $l_i \in \{\text{`h'}, \text{`t'}\}$ denotes the side of the relation (head or tail), $\tilde{\mathcal{N}}_{r_i}^{l_i} \subset \ents$ represents the set of entities connected to relation $r_i$ on the $l_i$ side. $r_s$ is a newly introduced relation in the semantic space, and the final node set of $\graph_{\it r}$ is equal to $(\rels \cup \{r_s\})$.

\section{Implementation Details}
\label{app:imd}

We introduce the statistics of pre-training KGs in Table \ref{table:ds-train}.
Following previous work \citep{REDGNN-WWW22, NBF-NIPS21}, we augment the triples in each $\graph$ with reverse and identity relations. The augmented triple set $\triples^+$ is defined as: $\triples^+ = \triples \cup \{(e_t, r^{-1}, e_h)|(e_h, r, e_t) \in \triples\} \cup \{(e, r^i, e)|e\in  \ents\}$, where the relation $r^{-1}$ is the reverse relation of a relation $r$, the relation $r^i$ refers to the identity relation, and the number of augmented triples is $|\triples^+| = 2|\triples|+|\ents|$.

We employ the ULTRA(3g)~\citep{ultra} model as the major baseline, utilizing the released checkpoint pre-trained on three knoglwedge graphs.
We evaluate OpenGraph~\citep{OpenGraph} and AnyGraph~\citep{AnyGraph} across diverse node-level datasets using their publicly released pre-trained model weights.
The metric results of GraphAny are referred to its official paper~\citep{GraphAny-Arxiv24}, and some node-level and graph-level results for graph prompt learning models are from the ProG work~\citep{ProG}.
Although there are some previous KG reasoning baselines \citep{NodePiece,DRUM-NIPS19,GraIL-ICML19} with no semantic features involving in, we ignore them in our more challenging generalized reasoning tasks.

We train \method with three commonly-used KG datasets, WN18RR \citep{WN18RR}, FB15k237 \citep{FB15k237}, and CodexM~\citep{Codex}, following the hyperparameter settings of ULTRA(3g). 
Concerns about potential relation leakage during pre-training can be ignored because neither our method nor ULTRA learns relation-specific parameters. 
Specifically, the CMP module follows NBFNet~\citep{NBF-NIPS21} with a non-parametric DistMult~\citep{DistMult} message function and a simplified PNA aggregation funcition~\citep{PNA-NIPS20}, which leverages only two sub-aggregations: MEAN and STD.
The number of layers $L$ for both CMP and SCMP is set to 6, with the hidden dimension configured at 64.
The relation encoder utilizes randomly initialized edge embeddings for $\mathcal{R}_{\it fund}$. In contrast, ${\it SCMP}(\cdot)$ initializes the embeddings of edge types using the relative relation embeddings $\mathbf{R}_q$. We suggest consulting the ULTRA paper \citep{ultra} for further details.

We introduce two common types of node semantic features in knowledge graphs for model training.
\begin{itemize}[noitemsep,topsep=0pt,parsep=0pt,partopsep=0pt,leftmargin=10pt]
    \item \textbf{Textual Embeddings} are vector representations of textual information associated with entities in a KG, typically generated using models like BERT~\citep{Bert} or Word2Vec~\citep{word2vec}. 
    Textual embeddings are broadly applicable across different KGs, as most KGs contain some form of textual metadata. 
    However, the richness and variety of text data across KGs—such as short labels or multilingual content—can introduce diversity in how these embeddings are utilized, requiring models to generalize across various linguistic and domain-specific contexts.
    \item \textbf{Ontology Features} refer to structured representations of entities within a formalized schema, such as a $|R|$-length vector that counts the relation types associated with each entity. 
    These features offer a global understanding of an entity’s role in the graph by capturing its relational context. 
    Common across various domains, they provide a simplified view of an entity’s interactions. 
    However, the diversity of relation types and their distribution can vary significantly across KGs, which affects how well these features generalize.
\end{itemize}

Hyperparameters are selected through grid search based on the metrics from the validation set. 
The similarity threshold $\delta$ and the number of neighbors k were not fine-tuned for individual datasets. Specifically, we set $\delta$=0.9 for all node/graph-level datasets. For the neighbor number k, all six node-level datasets share k=2, while graph-level datasets use either k=1 or k=3, depending on their domain. SCR maintains robust performance across datasets with a single task-level configuration, preserving its zero-shot capability.
All experiments are performed on Intel Xeon Gold 6238R CPU @ 2.20GHz and NVIDIA RTX A30 GPUs (four for pretraining and one for evaluation), and are implemented in Python using the PyTorch framework.
Our source code is implemented based on ULTRA\footnote{https://github.com/DeepGraphLearning/ULTRA}, which is available under the MIT License. 
All employed KG datasets are open and commonly used.

\section{Task-specific KG Structure}
\label{app:proof1}

\textbf{Node Classification Task:}
We first define the unified reasoning format from the view of node classification. Suppose $G = (V, E)$ is a graph in $\mathcal{D}$ whose nodes in $V$ must be classified by labels in a finite label set $\mathcal{L}$. We construct the new, heterogeneous graph $\widetilde{G} = (\widetilde{V}, \widetilde{E})$ as follows. 

Let $\widetilde{V} = V \,\cup\, \mathcal{L}$ be the set of all nodes in the new graph, where each $\ell \in \mathcal{L}$ is viewed as a distinct ``label node''.
Retain every original edge from $E$ in $\widetilde{G}$, so that if $(v,u)\in E$ in the original graph $G$, the same (possibly typed) edge is preserved in $\widetilde{G}$.
Furthermore, introduce a designated relation type $r_{\text{label}}$ connecting nodes $v \in V$ to label nodes $\ell \in \mathcal{L}$, i.e.:
\begin{align}
E_{\text{label}} 
= 
\bigl\{
\,&(v,\, r_{\text{label}},\, \ell) 
\;\mid\;
v \in V,\,
\ell \in \mathcal{L},\,\nonumber\\
&\text{and } v \text{ is assigned training label } \ell\text{ in } \mathcal{D}
\bigr\}.
\end{align}
Then let the edge set $\widetilde{E} = E \,\cup\, E_{\text{label}}$. This completes the construction of the heterogeneous graph $\widetilde{G} = (\widetilde{V}, \widetilde{E})$.

Node classification in $G$ bijectively map to link prediction in $\widetilde{G}$, observe that assigning a label $\ell$ to a node $v$ in the original problem becomes the presence of an edge $(v, r_{label}, \ell)$ in $\widetilde{G}$.
First, assume a labeling function $f:V \rightarrow 2^\mathcal{L}$ is given. Each instance \(v \mapsto \ell\) that appears in $f$ corresponds to including $(v,\, r_{\text{label}},\, \ell) \in E_{\text{label}}$. 
Hence the labeling of $G$ completely specifies the set of label-links in $\widetilde{G}$. 
Conversely, given a link $(v,\, r_{\text{label}},\, \ell) \in \widetilde{E}$, one uniquely recovers the statement that $v$ is labeled by $\ell$ in the original classification problem. 
This one-to-one correspondence implies that any function assigning labels to nodes in 
$G$ bijectively maps to a set of label-links in $\widetilde{G}$.

\textbf{Graph Classification Task:}
Let $\mathcal{D} = \{G_1, G_2, \dots, G_M\}$ be a collection of graphs, where each $G_i$ is to be assigned one or more labels from a finite label set $\mathcal{L} = \{\ell_1, \dots, \ell_K\}$. Suppose each graph $G_i$ in $\mathcal{D}$ has node set $V_i$ and edge set $E_i$. We aim to label each $G_i$ with one or more labels from $\mathcal{L}$. An extended graph $\widetilde{G}$ is constructed as follows.

\textit{Step 1:} Introduce a new \emph{graph node} $s_i$ for each graph $G_i$, which will represent the entire graph $G_i$ as a single entity in $\widetilde{G}$. 

\textit{Step 2:} Include in $\widetilde{G}$ all original nodes from each $G_i$. That is, take 
$V_{\mathrm{all}} \;=\; \bigcup_{i=1}^M V_i$, and add these re-indexed nodes to $\widetilde{G}$ along with their internal edges $E_{\mathrm{all}} \;=\; \bigcup_{i=1}^M E_i$.

\textit{Step 3:} For each $v \in V_i$, add an edge $(s_i,\,r_{\text{node-graph}},\,v)$ to indicate that $v$ is a member of the graph $G_i$. The relation $r_{\text{node-graph}}$ is a designated edge type (e.g., ``belongsToGraph'').

\textit{Step 4:} For labeling, add a node for each label $\ell \in \mathcal{L}$. Let these form the set of \emph{label nodes} in $\widetilde{G}$. To encode the classification of $G_i$ with label $\ell$, add an edge $(s_i,\,r_{\text{label}},\,\ell)$ whenever $G_i$ is assigned label $\ell$. Let $E_{\mathrm{label}}$ be the set of all such edges.

\textit{Step 5:} For edges between graph nodes, add an edge $\bigl(s_i,\,r_{\text{similar}},\,s_j)$ whenever the graph embedding vector of $G_i$ is close to that of $G_j$. Let $E_{\mathrm{similar}}$ be the set of all such edges.

Hence, the heterogeneous graph is merged as follows:
\begin{align}
\widetilde{G} \;=\; &\bigl(\, V_{\mathrm{all}} \,\cup\, \{s_1,\dots,s_M\} \,\cup\, \mathcal{L},\;\nonumber\\
&E_{\mathrm{all}}\,\cup\, E_{\mathrm{node-graph}} \,\cup\, E_{\mathrm{label}}\,\cup\, E_{\mathrm{similar}}\bigr).
\end{align}
Similar to the claims about node classification, labeling $G_i$ with $\ell$ in the original problem is exactly equivalent to the statement that $\bigl(s_i,\,r_{\text{label}},\,\ell\bigr)$ is an edge in $\widetilde{G}$.
Because every valid assignment $G_i \mapsto \ell$ corresponds bijectively to an edge $\bigl(s_i,\,r_{\text{label}},\,\ell\bigr)$.

\section{Expressive Power}
\label{app:proof}


We formally analyze the expressive power of \method by comparing it with ULTRA~\citep{ultra}. Following the theory of the Weisfeiler-Leman test, we measure expressivity via a method's ability to distinguish non-isomorphic subgraphs in knowledge graphs.
 
Firstly, we show that \method is at least as expressive as ULTRA. For any non-isomorphic graphs distinguishable by ULTRA, there exists a parameter configuration of \method that achieves identical distinguishability.

We establish this through architectural reduction. Let $\theta = (k, W_{\text{MLP}})$ denote \method's key hyperparameters where $k$ controls semantic neighborhood size and $W_{\text{MLP}}$ the MLP weights from Eq.~(7). When $\theta_0 = (0, \mathbf{0})$, \method reduces to ULTRA through three key simplifications:

\begin{enumerate}[leftmargin=*,nosep]
    \item The augmented relation graph $\mathcal{G}_r$ collapses to ULTRA's original structure by removing the semantic relation type $r_s$;
    \item The INIT function (Eq.~(5)) reduces to ULTRA's initialization by eliminating the $v_a$ term;
    \item The representation fusion becomes identity because $\text{MLP}(\mathbf{H}_g)=0$;
\end{enumerate}

Under $\theta_0$, the message passing dynamics of both architectures become isomorphic. Therefore, ULTRA constitutes a proper subspace of \method's parameter space.

Secondly, we indicate that there exists a class of non-isomorphic triples distinguishable by \method but not by ULTRA, provided that semantic features contain discriminative information beyond graph topology.

Consider two candidate entities $e_1, e_2 \in \mathcal{E}$ with identical topological signatures relative to query entity $e_q$:
\begin{equation}
    \forall p \in \Pi_{\text{path}}: f_{\text{ULTRA}}(e_q, p, e_1) = f_{\text{ULTRA}}(e_q, p, e_2)
\end{equation}
where $\Pi_{\text{path}}$ denotes relational paths and $f$ the path encoding function.

ULTRA cannot distinguish $(e_q, r_q, e_1)$ from $(e_q, r_q, e_2)$ since their topological embeddings coincide. However, if the semantic features satisfy:
\begin{equation}
    \min_{v \in \mathcal{N}_{r_q}(e_q)} \|\tilde{x}_{e_1} - \tilde{x}_v\|_2 \ll \min_{v \in \mathcal{N}_{r_q}(e_q)} \|\tilde{x}_{e_2} - \tilde{x}_v\|_2
\end{equation}
where $\mathcal{N}_{r_q}(e_q)$ are $r_q$-neighbors of $e_q$, then \method's semantic proximity measure induces divergent embeddings:
\begin{equation}
    \|h_{e_1}^{(L)} - h_{e_2}^{(L)}\|_2 \geq \gamma > 0
\end{equation}
The separation constant $\gamma$ persists through MLP fusion (Eq.~(7)) by the Lipschitz continuity of neural networks. Thus, \method distinguishes the triplets while ULTRA cannot.

This analysis reveals \method's enhanced expressiveness stems from its \textit{semantic-topological fusion} mechanism. By jointly optimizing structural and semantic proximity measures during pre-training, the model learns disentangled yet complementary representations that strictly subsume purely topological approaches like ULTRA.

\begin{figure}[!t]
\centering
\begin{subfigure}{1\textwidth}
    \centering
    \includegraphics[width=0.6\textwidth]{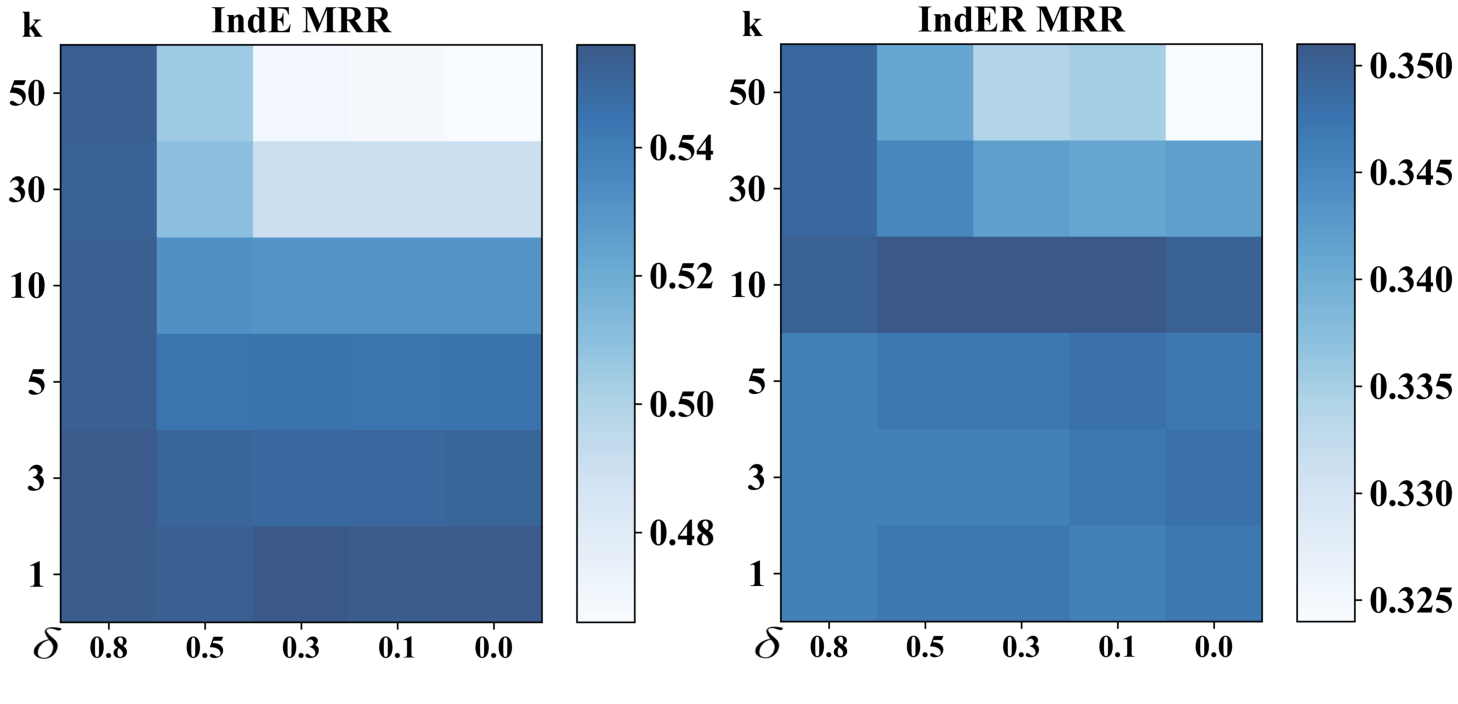}
    \label{fig:sensiMRR}
\end{subfigure}
\vspace{-2mm}
\begin{subfigure}{1\textwidth}
    \centering
\includegraphics[width=0.6\textwidth]{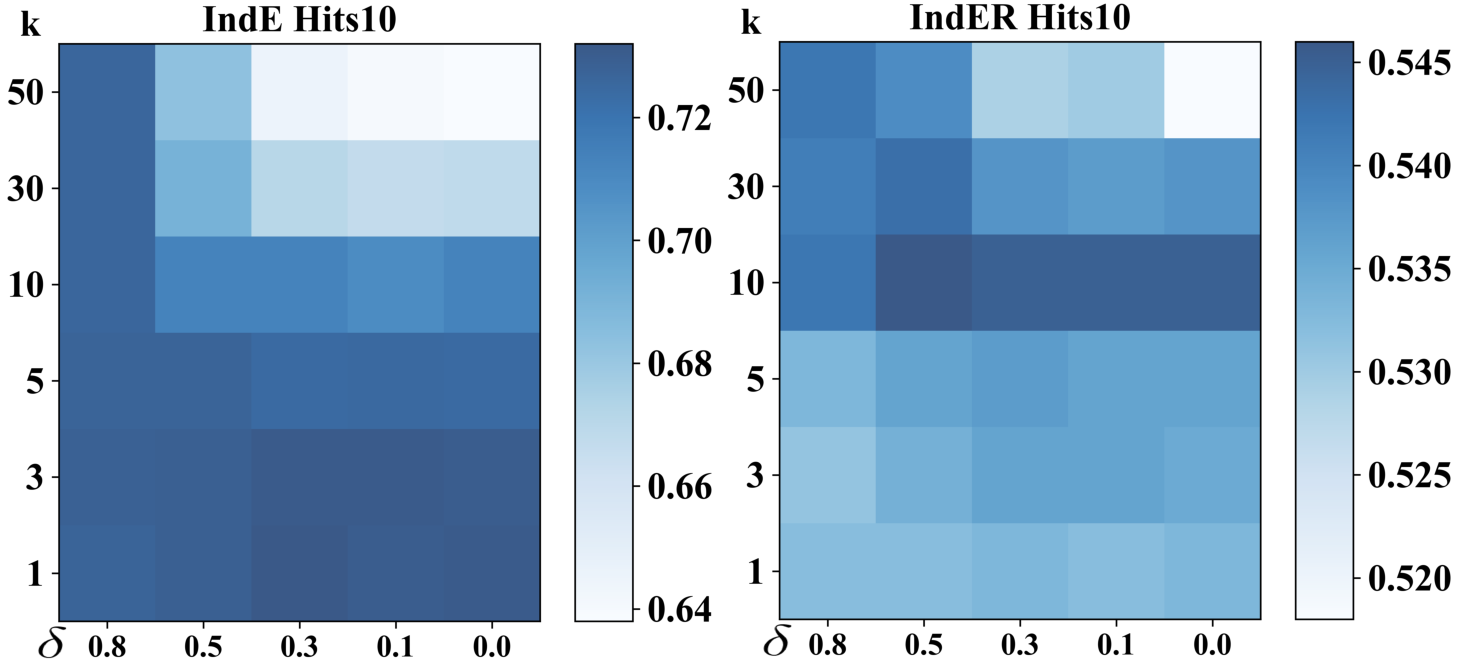}
    \label{fig:sensiHits10}
\end{subfigure}
\caption{Comparison of performance metrics under different hyperparameter settings ($k$ and $\delta$) for semantic neighbor selection. (a) Average MRR results. (b) Hits@10 results. Darker colors indicate higher values.}
\label{fig:combinedHeatmaps}
\end{figure}

\section{Hyperparameter Sensitivity (RQ5)}
\label{exp:RQ5}

Furthermore, we evaluate the impact of hyperparameters $k$ and $\delta$ on the selection of semantic neighbors. Here, $k$ specifies the number of neighbors, while $\delta$ determines the minimum similarity threshold.
As illustrated in Figure \ref{fig:combinedHeatmaps}, variations in these hyperparameters slightly affect prediction performance. Different choices of $k$ show similar performance across the IndE benchmarks, but for IndER datasets, $k = 10$ clearly outperforms other values. When $k>10$, the influence of $\delta$ becomes more pronounced, as a lower $\delta$ includes more dissimilar neighbors, negatively affecting model performance. 

\section{Assumption in \method}
\label{app:Assumption}
In Section \ref{sec:3.3}, we claim that the target node distinguishability assumption holds for the $INIT^2()$ function, if $\mathbf{r_q} \neq \mathbf{v_a}$ and neither of them contains zero entries.
This assumption usually holds because of the distinction between $\mathbf{r_q}$ (relation representations outputted by $CMP_{\phi}$) and $\mathbf{v_a}$ (trainable vector parameters) is inherently preserved through their distinct initialization and optimization mechanisms. Empirically, we conducted parameter analysis on our pre-trained SCR model and observed that: 
All $\mathbf{v_a}$ parameters maintained non-zero magnitudes (> 0.5511); $\mathbf{r_q}$ outputs on six datasets showed no exact zero entries or equality to $\mathbf{v_a}$. While theoretical equality is possible, it would require exact parameter convergence to zero or identical gradient updates—statistically implausible in practice.

\section{Complexity and Scalability Analysis}
\label{sec:complexity}
Given a knowledge graph $\graph = (\ents,\rels,\triples)$, we have that $|\ents|, |\rels|, |\triples|$ represent the size of entities, relation types, and triples, respectively. $\graph_r = (\rels,\rels_{fund},\triples_r)$ denotes the relation graph of $\graph$. $\features \in \mathbb{R}^{|\ents| \times d_0}$ is the input feature matrix, $d$ is the hidden dimension of the model, and $L$ is the number of layers in the model.
In the preprocessing stage, unifying semantic features requires a time complexity of $\mathcal{O}(|\ents| d d_0)$ for the SVD low-rank approximation. 
Our adoption of ``torch.svd\_lowrank(X, q=d, niter=2)'' leverages randomized truncated SVD, avoiding the $O(|E|^3)$ complexity of the full SVD.
Constructing relation graphs involves extracting the top K most similar neighbors for each entity, with a time complexity of $\mathcal{O}(|\ents|^2 (d + \log K))$, which simplifies to $\mathcal{O}(|\ents|^2 d)$ as $d \gg \log K$. Therefore, the overall complexity is $\mathcal{O}((d_0 + |\ents|)|\ents| d)$.

In terms of the CMP module, as shown in \citet{NBF-NIPS21}, the time complexity for a forward pass on $\graph$ is $\mathcal{O}(L(|\triples|d + |\ents|d^2))$ to compute one query reasoning. 
The runtimes of CMP and SCMP are comparable because SCMP's global encoding is shared across all queries, resulting in only a linear overhead. 
Combining the CMP calculations on $\graph_r$, the total complexity is $\mathcal{O}((|\triples|+|\triples_r|)Ld + (|\ents|+|\rels|)Ld^2)$.
Because one CMP-based reasoning calculates $|\ents|$ candidate triples at the same time, resulting in an amortized complexity is better than traditional relational message-passing models, such as RGCN~\citep{RGCN-ESWC18}, CompGCN~\citep{CompGCN-ICLR20}, and GraIL~\citep{GraIL-ICML19}.

Regarding scalability, \method exhibits comparable scalability and running time to ULTRA. 
To ease the concern, we conducted additional experiments in Appendix \ref{app:50KGs} on all transductive KG datasets mentioned in ULTRA, including large-scale datasets with over 100k triples, such as YAGO310.
We acknowledge that both methods face practical limitations when applied to KGs with millions or billions of triples. Specifically, the time cost of subgraph extraction and message passing becomes non-trivial compared to traditional embedding-based models. This limitation is inherent to subgraph-based inductive frameworks but does not preclude \method’s applicability to typical large-scale KGs.
We will prioritize this in future work. For large-scale KGs, recent acceleration techniques like TIGER~\citep{OurTIGER} (enabling efficient subgraph extraction for inductive reasoning on Freebase) are critical. While \method’s current implementation does not yet integrate these optimizations, its framework is compatible with such methods.

\section{Limitations}
\label{sec:limit}
Here, we discuss two limitations of this work. First, the training data for \method is confined to three popular KGs. There is room for improvement by training with more diverse graphs, particularly on challenging tasks like those involving heterophilic graphs. 
Second, the scalability of \method on larger-scale graphs remains to be verified in future work. The expectation is positive, given the recent efforts focused on accelerating KG reasoning through system and algorithmic optimizations~\citep{Adaprop,ANet,OurTIGER}.
Third, this paper presents work whose goal is to advance the field of Machine Learning. There are many potential societal consequences of our work, none of which we feel must be specifically highlighted here.
In this work, the LLM is used only for writing and editing.

\section{Evaluation on More KG Datasets}
\label{app:50KGs}
We rigorously test on all 54 datasets and the 3 pre-training graphs referenced in ULTRA’s framework~\citep{ultra}. 
The results shown in Table \ref{tab:50kgs} confirm that \method outperforms ULTRA across the majority of metrics when evaluated under identical conditions (PyG-based implementation and pre-training data). 

\begin{table}[!t]
\caption{Per-dataset and average performance of ULTRA and \method on 54 KG datasets.}
\label{tab:50kgs}
\centering
\setlength{\tabcolsep}{1.2mm}
\small
\begin{tabular}{l|cccc}\toprule
\multirow{2}{*}{KG Datasets} & \multicolumn{2}{c}{ULTRA(3g)} & \multicolumn{2}{c}{SCR(3g)} \\
                             & MRR       & {Hits@10}     & MRR      & {Hits@10}    \\
\hline
WikiTopicsMT1:tax            & 0.242     & 0.305             & 0.182    & 0.312            \\
WikiTopicsMT1:health         & 0.279     & 0.332             & 0.265    & 0.410            \\
WikiTopicsMT2:org            & 0.083     & 0.145             & 0.078    & 0.139            \\
WikiTopicsMT2:sci            & 0.258     & 0.348             & 0.245    & 0.354            \\
WikiTopicsMT3:art            & 0.251     & 0.414             & 0.244    & 0.407            \\
WikiTopicsMT3:infra          & 0.622     & 0.779             & 0.635    & 0.781            \\
WikiTopicsMT4:sci            & 0.293     & 0.455             & 0.256    & 0.461            \\
WikiTopicsMT4:health         & 0.557     & 0.707             & 0.615    & 0.753            \\
Metafam                      & 0.330     & 0.821             & 0.246    & 0.560            \\
FBNELL                       & 0.473     & 0.653             & 0.480    & 0.676            \\
\hline
ILPC2022:small               & 0.296     & 0.441             & 0.285    & 0.443            \\
ILPC2022:large               & 0.297     & 0.423             & 0.285    & 0.419            \\
HM:1k                        & 0.079     & 0.150             & 0.055    & 0.097            \\
HM:3k                        & 0.063     & 0.120             & 0.047    & 0.083            \\
HM:5k                        & 0.055     & 0.101             & 0.041    & 0.072            \\
HM:indigo                    & 0.436     & 0.649             & 0.425    & 0.631            \\
\hline
YAGO310                      & 0.480     & 0.658             & 0.488    & 0.666            \\
NELL995                      & 0.437     & 0.575             & 0.456    & 0.608            \\
CoDExSmall                   & 0.472     & 0.668             & 0.436    & 0.653            \\
CoDExLarge                   & 0.333     & 0.461             & 0.329    & 0.458            \\
Hetionet                     & 0.261     & 0.382             & 0.289    & 0.402            \\
ConceptNet100k               & 0.061     & 0.117             & 0.115    & 0.218            \\
DBpedia100k                  & 0.397     & 0.565             & 0.401    & 0.573            \\
AristoV4                     & 0.183     & 0.262             & 0.227    & 0.349            \\
WDsinger                     & 0.370     & 0.488             & 0.371    & 0.498            \\
NELL23k                      & 0.241     & 0.406             & 0.234    & 0.402            \\
FB15k237\_10                 & 0.159     & 0.272             & 0.155    & 0.265            \\
FB15k237\_20                 & 0.183     & 0.309             & 0.179    & 0.305            \\
FB15k237\_50                 & 0.230     & 0.396             & 0.222    & 0.389            \\
\hline
FB15k237                     & 0.369     & 0.562             & 0.344    & 0.533            \\
WN18RR                       & 0.369     & 0.533             & 0.444    & 0.571            \\
CoDExMedium                  & 0.374     & 0.527             & 0.350    & 0.498            \\
\hline
Inductive e,r (23 graphs)    & 0.342     & 0.510             & 0.338    & 0.516            \\
Inductive e (18 graphs)      & 0.416     & 0.568             & 0.433    & 0.582            \\
Transductive (13 graphs)     & 0.293     & 0.428             & 0.300    & 0.445            \\
Total AVG (54 graphs)        & 0.355     & 0.510             & 0.361    & 0.521            \\
Pretraining (3 graphs)       & 0.371     & 0.541             & 0.379    & 0.534           \\
\bottomrule
\end{tabular}
\vspace{-3mm}
\end{table}

\begin{table*}[!t]
\caption{Inductive KG datasets used in the experiments. "Triples" refers to the number of edges in the graph used for training, validation, or testing. "Valid" and "Test" refer to the triples that need to be predicted in the validation and test sets, respectively, within the corresponding graphs.}
\label{table:ds-test}
\setlength{\tabcolsep}{1.2mm}
\scriptsize
\begin{tabular}{clcccccccccccc}\toprule
\multirow{2}{*}{Group} &\multirow{2}{*}{Dataset} &\multicolumn{3}{c}{Training Graph} &\multicolumn{4}{c}{Validation Graph} &\multicolumn{4}{c}{Test Graph} \\ \cmidrule(l){3-5} \cmidrule(l){6-9} \cmidrule(l){10-13}
& &Entities &Rels &Triples &Entities &Rels &Triples &Valid &Entities &Rels &Triples &Test \\\midrule
\multirow{4}{*}{IndE(FB)} & FB:v1~\citep{GraIL-ICML19} &1594 &180 &4245 &1594 &180 &4245 &489 &1093 &180 &1993 &411 \\
& FB:v2~\citep{GraIL-ICML19} &2608 &200 &9739 &2608 &200 &9739 &1166 &1660 &200 &4145 &947 \\
& FB:v3~\citep{GraIL-ICML19} &3668 &215 &17986 &3668 &215 &17986 &2194 &2501 &215 &7406 &1731 \\
& FB:v4~\citep{GraIL-ICML19} &4707 &219 &27203 &4707 &219 &27203 &3352 &3051 &219 &11714 &2840 \\
\multirow{4}{*}{IndE(WN)} & WN:v1~\citep{GraIL-ICML19} &2746 &9 &5410 &2746 &9 &5410 &630 &922 &9 &1618 &373 \\
& WN:v2~\citep{GraIL-ICML19} &6954 &10 &15262 &6954 &10 &15262 &1838 &2757 &10 &4011 &852  \\
& WN:v3~\citep{GraIL-ICML19} &12078 &11 &25901 &12078 &11 &25901 &3097 &5084 &11 &6327 &1143 \\
& WN:v4~\citep{GraIL-ICML19} &3861 &9 &7940 &3861 &9 &7940 &934 &7084 &9 &12334 &2823\\
\multirow{4}{*}{IndE(NL)} & NL:v1~\citep{GraIL-ICML19} &3103 &14 &4687 &3103 &14 &4687 &414 &225 &14 &833 &201 \\
& NL:v2~\citep{GraIL-ICML19} &2564 &88 &8219 &2564 &88 &8219 &922 &2086 &88 &4586 &935 \\
& NL:v3~\citep{GraIL-ICML19} &4647 &142 &16393 &4647 &142 &16393 &1851 &3566 &142 &8048 &1620 \\
& NL:v4~\citep{GraIL-ICML19} &2092 &76 &7546 &2092 &76 &7546 &876 &2795 &76 &7073 &1447\\
\midrule
\multirow{4}{*}{IndER(FB)} & FB-25~\citep{ingram} &5190 &163 &91571 &4097 &216 &17147 &5716 &4097 &216 &17147 &5716 \\
& FB-50~\citep{ingram} &5190 &153 &85375 &4445 &205 &11636 &3879 &4445 &205 &11636 &3879 \\
& FB-75~\citep{ingram} &4659 &134 &62809 &2792 &186 &9316 &3106 &2792 &186 &9316 &3106  \\
& FB-100~\citep{ingram} &4659 &134 &62809 &2624 &77 &6987 &2329 &2624 &77 &6987 &2329 \\
\multirow{4}{*}{IndER(WK)} & WK-25~\citep{ingram} &12659 &47 &41873 &3228 &74 &3391 &1130 &3228 &74 &3391 &1131 \\
& WK-50~\citep{ingram} &12022 &72 &82481 &9328 &93 &9672 &3224 &9328 &93 &9672 &3225  \\
& WK-75~\citep{ingram} &6853 &52 &28741 &2722 &65 &3430 &1143 &2722 &65 &3430 &1144 \\
& WK-100~\citep{ingram} &9784 &67 &49875 &12136 &37 &13487 &4496 &12136 &37 &13487 &4496  \\
\multirow{4}{*}{IndER(NL)} & NL-25~\citep{ingram} &4396 &106 &17578 &2146 &120 &2230 &743 &2146 &120 &2230 &744 \\
& NL-50~\citep{ingram} &4396 &106 &17578 &2335 &119 &2576 &859 &2335 &119 &2576 &859  \\
& NL-75~\citep{ingram} &2607 &96 &11058 &1578 &116 &1818 &606 &1578 &116 &1818 &607 \\
& NL-100~\citep{ingram} &1258 &55 &7832 &1709 &53 &2378 &793 &1709 &53 &2378 &793 \\
\bottomrule
\end{tabular}
\vspace{-3mm}
\end{table*}

\begin{table*}
\centering
\caption{Statistics of node/graph classification datasets.}
\scriptsize
\setlength{\tabcolsep}{1mm}
\label{table:ds-class}
\begin{tabular}{p{0.18\textwidth}<{\centering}cccp{0.12\textwidth}<{\centering}p{0.08\textwidth}<{\centering}p{0.25\textwidth}<{\centering}}
\cmidrule[1.2pt]{1-7}
Dataset & Graphs & Nodes & Edges &Feature Dims &Classes & Node-level Task \\
\cmidrule{1-7}
Cora   & 1    & 2,708      & 5,429      & 1,433         & 7   &Homophilic Node Classification \\
CiteSeer  & 1  & 3,327      & 9,104      & 3,703         & 6    &Homophilic Node Classification\\
Pubmed   & 1    &19,717      & 88,648     & 500           & 3    &Homophilic Node Classification\\
Actor  & 1    & 7600       & 30,019       & 932       & 5      & Heterophilic Node Classification\\
Wisconsin & 1  & 251       & 515       & 1,703       & 5       & Heterophilic Node Classification\\
Texas & 1  & 183       & 325       & 1703       & 5    & Heterophilic Node Classification    \\
\cmidrule{1-7}
Dataset & Graphs   &Avg.nodes &Avg.edges &Feature Dims &Classes & Graph-level Task \\
\cmidrule{1-7}
IMDB-BINARY  & 1,000    & 19.8       & 96.53       & 0     & 2          & Social Network Classification\\
COLLAB  & 5,000    & 74.5       & 2457.8       & 0        & 3         & Social Network Classification\\
PROTEINS & 1,113 & 39.1 & 72.8 & 3  & 2  &Protein Graph Classification\\
ENZYMES & 600  & 32.6 & 62.1 & 3 & 6 & Protein Graph Classification\\
DD & 1,178  & 284.1 & 715.7 & 89 & 2 & Protein Graph Classification\\
MUTAG     & 188         & 17.9       & 19.8       & 7        & 2        &Small Molecule Classification \\
COX2 & 467  & 41.2 & 43.5 & 3 & 2  & Small Molecule Classification \\
BZR & 405  & 35.8 & 38.4 & 3 & 2 & Small Molecule Classification \\
\cmidrule[1.2pt]{1-7}
\end{tabular}
\end{table*}

\section{Related Work}
\label{sec:relatedwork}
\subsection{Knowledge Graph Reasoning}

Traditional \emph{transductive} KG reasoning models, such as TransE~\citep{TransE}, DistMult~\citep{DistMult}, RotatE~\citep{RotatE}, RGCN~\citep{RGCN-ESWC18}, and CompGCN~\citep{CompGCN-ICLR20}, represent entities and relations within a knowledge graph using continuous vector embeddings~\citep{2017Survey,OurHaLE}. 
These models, however, assume that all entities and relations in the KG are known beforehand, which limits their ability to generalize to unseen entities within the same graph or across different KGs~\citep{OurMulDE,GoogleAttH}. 
In contrast, \emph{inductive} KG reasoning approaches~\citep{NBF-NIPS21} address this limitation by enabling generalization to KGs with previously unseen entities or relations. 
Most existing inductive methods~\citep{Cycle-ICML22,InductivE-IJCNN21,IKR-JWWW23,TACT-AAAI21} employ query-conditional MPNNs to generate ``relative'' entity embeddings by extracting local structural features from a subgraph induced by the query entity. 
GraIL~\citep{GraIL-ICML19}, for example, extracts an enclosing subgraph between the query entity and each candidate entity, but this approach suffers from high computational costs. 
Other models, such as NBFNet~\citep{NBF-NIPS21} and RED-GNN~\citep{REDGNN-WWW22}, propagate query features through the $L$-hop neighborhood subgraph of the query entity. 
To improve computational efficiency, recent works have focused on optimizing algorithms, including path-pruning techniques during the GNN propagation process~\citep{Adaprop,ANet,OurGraPE}.
In the direction of building a foundation model for KG reasoning, ULTRA~\citep{ultra} utilizes four basic interaction types of the KG relational structure to perform inductive reasoning on entirely novel KGs. 
Building on ULTRA, ProLINK~\citep{OurProLINK} harnesses the power of large language models (LLMs) to enhance reasoning performance for few-shot relation types on low-resource KGs. 
Nevertheless, these inductive methods face challenges in generalizing to a wide variety of graph tasks and feature spaces, as their reasoning capabilities remain primarily confined to topological structures.


\subsection{Graph Foundation Models}

Graph foundation models are increasingly recognized for their ability to manage diverse graph-structured data across various tasks. 
Traditionally, model fine-tuning offers a simple method to adapt these models to downstream tasks. 
However, significant discrepancies between tasks can lead to negative transfer and catastrophic forgetting~\citep{rosenstein2005transfer}. 
An alternative to fine-tuning is graph prompt learning, which reformulates input graph data to better align with the pretext task~\citep{chen2023ultradp, tan2023virtual, gong2023prompt}.
In this context, GPF utilizes a prompt token by adding supplementary features to the base graph. 
Building on this, GPF-plus~\citep{gpfplus2023} trains multiple independent basis vectors and integrates them through attentive aggregation facilitated by several learnable linear projections.
GPPT~\citep{sun2022gppt} introduces graph prompts as additional tokens comprising task-specific and structural elements, aiding in node tasks and link prediction pretext alignment. 
Gprompt~\citep{liu2023graphprompt} incorporates prompt vectors into graph pooling through element-wise multiplication.
Other research considers graph prompts as additional graphs~\citep{ge2023enhancing, huang2023prodigy}. The All-in-one model~\citep{sun2023all}, for instance, integrates token graphs as prompts within the original graph, linking tokens directly with the original graph elements. 
Although these graph prompt learning methods achieve good performance in various graph tasks, they require an additional learning phase for task-specific parameters.

Recently, several foundational models for specific graph tasks have been proposed to adapt to diverse unseen data without model tuning.
ULTRA~\citep{ultra} and KG-ICL~\citep{KG-ICL} are pre-trained on multiple KGs to obtain the capability of reasoning on new KGs. ProLINK~\citep{OurProLINK} and TRIX~\citep{TRIX} further expand on ULTRA with prompt graphs from LLMs and iterative updates of relation/entity embeddings.
GraphAny~\citep{GraphAny-Arxiv24} addresses inductive node classification by formulating inference as an analytical solution to a linear GNN architecture, while an attention module fuses predictions from multiple models, ensuring scalability. 
Models like InstructGLM~\citep{InstructGLM} and HiGPT~\citep{HiGPT} leverage large language models (LLMs), using natural language prompts to guide graph learning and handling heterogeneous graphs without downstream fine-tuning, broadening the applicability of foundation models to diverse graph tasks.
Building a general graph foundation model is not trivial; the major challenges to overcome are related to structural and feature heterogeneity.
OpenGraph~\citep{OpenGraph} proposes a zero-shot graph learning framework with a unified graph tokenizer and a scalable graph transformer, allowing the model to handle unseen graph data, aided by LLM-based data augmentation. 
AnyGraph~\citep{AnyGraph} extends this by addressing structural and feature heterogeneity through a Graph Mixture-of-Experts architecture, supporting fast adaptation and scaling efficiently. 
Research on Graph Foundation Models is still in its early stages, and current methods often struggle to match the competitive performance of task-specific supervised methods~\citep{GFM_HKPU_arxiv2503,GFMSurvey-arxiv23}.

\section{Additional Experimental Results}
\label{app:addexp}

\begin{table*}[!t]
\centering
\setlength\tabcolsep{3pt}
\caption{Accuracy results on node classification datasets where 90\% samples are divided into the test set.}
\small
\label{tab:node2}
\begin{adjustbox}{max width=\textwidth}
\begin{tabular}{c|l|lll|lll}
\toprule
Learning Paradigm & Methods & Cora* & Citeseer* & Pubmed* & Wisconsin* & Texas* & Actor*\\
\midrule
\multirow{2}{*}{\makecell{One-Shot Training}} & GCN & 26.56\scriptsize{$\pm$5.55} & 21.78\scriptsize{$\pm$7.32} & 39.37\scriptsize{$\pm$16.34} & 41.60\scriptsize{$\pm$3.10} & 37.97\scriptsize{$\pm$5.80} & 20.57\scriptsize{$\pm$4.47} \\
& Pre-train \& Fine-Tune & 40.40\scriptsize{$\pm$4.66} & 35.05\scriptsize{$\pm$4.37} & 46.74\scriptsize{$\pm$14.09} & 40.69\scriptsize{$\pm$4.13} &  47.66\scriptsize{$\pm$2.37} &  20.74\scriptsize{$\pm$4.12} \\
\midrule
\multirow{5}{*}{\makecell{Graph Pre-Training\\One-Shot Tuning}} & GPPT & 43.15\scriptsize{$\pm$9.44} & 37.26\scriptsize{$\pm$6.17} & 48.31\scriptsize{$\pm$17.72} & 30.40\scriptsize{$\pm$6.81} & 31.81\scriptsize{$\pm$15.33} & 22.58\scriptsize{$\pm$1.97}  \\
& All-in-one & 52.39\scriptsize{$\pm$10.17} & 40.41\scriptsize{$\pm$2.80} & 45.17\scriptsize{$\pm$6.45} & 78.24\scriptsize{$\pm$16.68} & 65.49\scriptsize{$\pm$7.06} & 24.61\scriptsize{$\pm$2.80}  \\
& Gprompt & 56.66\scriptsize{$\pm$11.22} & 53.21\scriptsize{$\pm$10.94} & 39.74\scriptsize{$\pm$15.35} & 83.80\scriptsize{$\pm$2.44} & 33.25\scriptsize{$\pm$40.11} & 25.26\scriptsize{$\pm$1.10} \\
& GPF & 38.57\scriptsize{$\pm$5.41} & 31.16\scriptsize{$\pm$8.05} & 49.99\scriptsize{$\pm$8.86} & 88.67\scriptsize{$\pm$5.78} & 87.40\scriptsize{$\pm$3.40} & 28.70\scriptsize{$\pm$3.35}\\
& GPF-plus & 55.77\scriptsize{$\pm$10.30} & 59.67\scriptsize{$\pm$11.87} & 46.64\scriptsize{$\pm$18.97} & \textbf{91.03\scriptsize{$\pm$4.11}} & \textbf{95.83\scriptsize{$\pm$4.19}} & 29.32\scriptsize{$\pm$8.56} \\
\midrule
\multirow{3}{*}{\makecell{KG Pre-Training\\No Tuning}} &  \textbf{\method(3g)} & 76.18\scriptsize{$\pm$0.09} & 50.40\scriptsize{$\pm$0.08} & 72.76\scriptsize{$\pm$0.14} & 46.67\scriptsize{$\pm$0.28} & 54.15\scriptsize{$\pm$0.24} & 23.52\scriptsize{$\pm$0.14} \\
& \method-20\% & 58.81\scriptsize{$\pm$2.74} & 36.62\scriptsize{$\pm$1.62} & 67.94\scriptsize{$\pm$0.56} & 45.42\scriptsize{$\pm$1.47} & 52.68\scriptsize{$\pm$0.83} & 21.81\scriptsize{$\pm$0.96} \\
& \method-5 & 56.80\scriptsize{$\pm$3.55} & 32.16\scriptsize{$\pm$3.85} & 51.63\scriptsize{$\pm$5.46} & 45.60\scriptsize{$\pm$2.06} & 56.59\scriptsize{$\pm$0.71} & 20.68\scriptsize{$\pm$1.09} \\
\bottomrule
\end{tabular}
\end{adjustbox}
\end{table*}

\begin{table*}[!t]
\centering
\setlength\tabcolsep{3pt}
\caption{F1 results on node classification datasets where 90\% samples are divided into the test set.}
\small
\label{tab:node3}
\begin{adjustbox}{max width=\textwidth}
\begin{tabular}{c|l|lll|lll}
\toprule
Learning Paradigm & Methods & Cora* & Citeseer* & Pubmed* & Wisconsin* & Texas* & Actor*\\
\midrule
\multirow{2}{*}{\makecell{One-Shot Training}} & GCN & 16.60\scriptsize{$\pm$2.54} & 10.81\scriptsize{$\pm$4.90} & 37.23\scriptsize{$\pm$15.48} & 26.34\scriptsize{$\pm$4.01} & 24.05\scriptsize{$\pm$5.12} & 11.56\scriptsize{$\pm$3.08} \\
& Pre-train \& Fine-Tune & 35.92\scriptsize{$\pm$4.06} & 30.78\scriptsize{$\pm$3.91} & 41.03\scriptsize{$\pm$13.36} & 27.43\scriptsize{$\pm$4.47} & 29.53\scriptsize{$\pm$6.44} &  15.91\scriptsize{$\pm$0.98} \\
\midrule
\multirow{5}{*}{\makecell{Graph Pre-Training\\One-Shot Tuning}} & GPPT	& 38.99\scriptsize{$\pm$8.32}	& 33.00\scriptsize{$\pm$6.49}	& 46.43\scriptsize{$\pm$16.73}	& 23.74\scriptsize{$\pm$5.95}	& 25.64\scriptsize{$\pm$8.12}	& 19.62\scriptsize{$\pm$0.56} \\
& All-in-one	& 46.58\scriptsize{$\pm$8.42}	& 30.20\scriptsize{$\pm$4.44}	& 38.05\scriptsize{$\pm$6.24}	& 67.68\scriptsize{$\pm$10.36}	& 43.37\scriptsize{$\pm$16.01}	& 16.05\scriptsize{$\pm$3.88}\\
& GPrompt	& 46.28\scriptsize{$\pm$8.46}	& 49.65\scriptsize{$\pm$11.42}	& 39.46\scriptsize{$\pm$15.97}	& 77.03\scriptsize{$\pm$6.40}	& 29.20\scriptsize{$\pm$35.62}	& 22.00\scriptsize{$\pm$1.74}\\
& GPF	& 23.79\scriptsize{$\pm$5.49}	& 18.63\scriptsize{$\pm$7.34}	& 45.36\scriptsize{$\pm$15.88}	& 82.97\scriptsize{$\pm$6.10}	& 78.43\scriptsize{$\pm$9.49}	& 31.69\scriptsize{$\pm$5.47}\\
& GPF-plus	& 53.28\scriptsize{$\pm$11.46}	& 56.22\scriptsize{$\pm$13.99}	& 42.38\scriptsize{$\pm$19.01}	& 85.24\scriptsize{$\pm$5.45}	& 86.22\scriptsize{$\pm$10.29}	& 24.56\scriptsize{$\pm$8.79} \\
\midrule
\multirow{3}{*}{\makecell{KG Pre-Training\\No Tuning}} &  \textbf{\method(3g)} & 73.06\scriptsize{$\pm$0.12} & 50.02\scriptsize{$\pm$0.05} & 70.79\scriptsize{$\pm$0.18} & 17.82\scriptsize{$\pm$0.05} & 17.51\scriptsize{$\pm$0.49} & 19.64\scriptsize{$\pm$0.15} \\
& \method-20\% & 55.81\scriptsize{$\pm$3.87} & 36.27\scriptsize{$\pm$3.43} & 64.92\scriptsize{$\pm$0.90} & 17.14\scriptsize{$\pm$4.06} & 16.05\scriptsize{$\pm$2.91} & 19.14\scriptsize{$\pm$0.52} \\
& \method-5 & 55.56\scriptsize{$\pm$2.29} & 28.10\scriptsize{$\pm$3.12} & 47.19\scriptsize{$\pm$6.23} & 27.41\scriptsize{$\pm$2.26} & 23.73\scriptsize{$\pm$0.37} & 18.74\scriptsize{$\pm$0.64} \\
\bottomrule
\end{tabular}
\end{adjustbox}
\end{table*}

\begin{table*}[!t]
\centering
\setlength\tabcolsep{5pt}
\small
\caption{F1 performance on node classification datasets. }
\begin{adjustbox}{max width=\textwidth}
\begin{tabular}{ll|lll|lll}
\toprule
 & Methods & Cora & Citeseer & Pubmed & Wisconsin & Texas & Actor\\
\midrule
\multirow{2}{*}{\makecell{One}} & GCN 	& 16.60\scriptsize{$\pm$2.54}	& 10.81\scriptsize{$\pm$4.90}	& 37.23\scriptsize{$\pm$15.48}	& 26.34\scriptsize{$\pm$4.01}	& 24.05\scriptsize{$\pm$5.12}	& 11.56\scriptsize{$\pm$3.08} \\
& Pre-train \& Fine-tune	& 35.92\scriptsize{$\pm$4.06}	& 30.78\scriptsize{$\pm$3.91}	& 41.03\scriptsize{$\pm$13.36}	& 26.74\scriptsize{$\pm$3.28}	& 29.53\scriptsize{$\pm$6.44}	& 15.91\scriptsize{$\pm$0.98} \\
\midrule
\multirow{3}{*}{\makecell{Graph Pre-Training\\No Tuning}} & OpenGraph	& 79.85\scriptsize{$\pm$0.71}	& 67.52\scriptsize{$\pm$0.75}	& 77.74\scriptsize{$\pm$1.65}	& 15.45\scriptsize{$\pm$3.00}	& 17.78\scriptsize{$\pm$5.07}	& 9.84\scriptsize{$\pm$2.66}\\
& AnyGraph (Link1)	& 60.5\scriptsize{$\pm$5.28}	& 49.81\scriptsize{$\pm$5.18}	& 58.44\scriptsize{$\pm$4.28}	& 1.33\scriptsize{$\pm$0.34}	& 0.40\scriptsize{$\pm$0.19}	& 4.98\scriptsize{$\pm$0.27} \\
& AnyGraph (Link2)	& 68.5\scriptsize{$\pm$3.16}	& 43.47\scriptsize{$\pm$3.34}	& 75.91\scriptsize{$\pm$1.54}	& 1.27\scriptsize{$\pm$0.22}	& 0.68\scriptsize{$\pm$0.47}	& 4.93\scriptsize{$\pm$0.31} \\
\midrule
\multirow{2}{*}{\makecell{KG Pre-Training\\No Tuning}} & ULTRA(3g) & 78.40\scriptsize{$\pm$0.00} & 64.68\scriptsize{$\pm$0.00} & 76.15\scriptsize{$\pm$0.00} & 25.71\scriptsize{$\pm$0.00} & 19.81\scriptsize{$\pm$0.00} & 14.62\scriptsize{$\pm$0.00} \\
& \method(3g) & 80.92\scriptsize{$\pm$0.61} & 69.24\scriptsize{$\pm$1.10} & 77.91\scriptsize{$\pm$1.31}
& 29.03\scriptsize{$\pm$3.66}	& 28.73\scriptsize{$\pm$1.59}	& 20.29\scriptsize{$\pm$0.41} \\
\midrule
\multirow{2}{*}{\makecell{Few-Shot Labeling}} & \method-20\% & 71.98\scriptsize{$\pm$2.95} & 53.48\scriptsize{$\pm$2.34} & 69.09\scriptsize{$\pm$0.34} & 23.62\scriptsize{$\pm$1.30} & 26.74\scriptsize{$\pm$7.10} & 19.45\scriptsize{$\pm$0.57} \\
& \method-5 & 54.08\scriptsize{$\pm$1.90} & 32.35\scriptsize{$\pm$2.71} & 47.19\scriptsize{$\pm$4.45} & 19.76\scriptsize{$\pm$2.84} & 37.29\scriptsize{$\pm$6.94} & 18.19\scriptsize{$\pm$0.91} \\
\bottomrule
\end{tabular}
\end{adjustbox}
\end{table*}

\begin{table*}[!t]
\centering
\small
\caption{F1 performance on graph classification datasets.}
\begin{adjustbox}{max width=\textwidth}
\begin{tabular}{ll|cccccccc}
\toprule
& Methods & IMDB-B & COLLAB & PROTEINS & MUTAG & ENZYMES & COX2 & BZR & DD \\
\midrule
\multirow{2}{*}{\makecell{One-Shot Training}} & GCN 	& 54.62\scriptsize{$\pm$1.12}	& 41.10\scriptsize{$\pm$0.39}	& 46.69\scriptsize{$\pm$10.82}	& 63.47\scriptsize{$\pm$6.36}	& 15.25\scriptsize{$\pm$3.96}	& 22.78\scriptsize{$\pm$10.69}	& 23.71\scriptsize{$\pm$8.23}	& 44.74\scriptsize{$\pm$4.23} \\
& Pre-train \& Fine-tune	& 55.24\scriptsize{$\pm$1.07}	& 41.71\scriptsize{$\pm$0.17}	& 59.73\scriptsize{$\pm$1.34}	& 63.70\scriptsize{$\pm$5.32}	& 19.17\scriptsize{$\pm$3.42}	& 45.06\scriptsize{$\pm$1.93}	& 33.12\scriptsize{$\pm$7.45}	& 48.68\scriptsize{$\pm$6.42} \\
\midrule
\multirow{5}{*}{\makecell{Graph Pre-Training\\One-Shot Tuning}} &GPPT	& 44.16\scriptsize{$\pm$6.70}	& 42.87\scriptsize{$\pm$7.70}	& 47.07\scriptsize{$\pm$11.95}	& 53.15\scriptsize{$\pm$16.82}	& 19.87\scriptsize{$\pm$2.99}	& 44.68\scriptsize{$\pm$1.17}	& 49.40\scriptsize{$\pm$8.41}	& 51.50\scriptsize{$\pm$6.54} \\
& All-in-one	& 56.88\scriptsize{$\pm$0.80}	& 47.78\scriptsize{$\pm$0.10}	& 64.68\scriptsize{$\pm$5.35}	& 78.57\scriptsize{$\pm$4.92}	& 19.66\scriptsize{$\pm$3.11}	& 49.62\scriptsize{$\pm$10.42}	& 62.11\scriptsize{$\pm$7.06}	& 56.70\scriptsize{$\pm$1.89} \\
& GPrompt	& 52.10\scriptsize{$\pm$13.61}	& 43.35\scriptsize{$\pm$10.75}	& 58.30\scriptsize{$\pm$10.88}	& 71.38\scriptsize{$\pm$3.64}	& 19.52\scriptsize{$\pm$3.36}	& 46.26\scriptsize{$\pm$5.14}	& 44.81\scriptsize{$\pm$6.73}	& 52.80\scriptsize{$\pm$3.60} \\
& GPF	& 56.22\scriptsize{$\pm$6.17}	& 38.14\scriptsize{$\pm$0.44}	& 57.01\scriptsize{$\pm$5.79}	& 63.90\scriptsize{$\pm$4.05}	& 17.34\scriptsize{$\pm$2.45}	& 43.08\scriptsize{$\pm$4.88}	& 48.83\scriptsize{$\pm$5.30}	& 48.52\scriptsize{$\pm$7.11}\\
& GPF-plus	& 55.55\scriptsize{$\pm$2.03}	& 41.24\scriptsize{$\pm$0.31}	& 57.58\scriptsize{$\pm$7.28}	& 63.20\scriptsize{$\pm$5.31}	& 18.39\scriptsize{$\pm$2.76}	& 30.90\scriptsize{$\pm$11.56}	& 46.57\scriptsize{$\pm$4.62}	& 46.24\scriptsize{$\pm$4.86} \\
\midrule
\multirow{2}{*}{\makecell{KG Pre-Training\\No Tuning}} & ULTRA(3g) & 38.87\scriptsize{$\pm$0.00} & 23.04\scriptsize{$\pm$0.00} & 37.48\scriptsize{$\pm$0.00} & 38.78\scriptsize{$\pm$0.00} & 5.84\scriptsize{$\pm$0.00} & 43.74\scriptsize{$\pm$0.00} & 44.23\scriptsize{$\pm$0.00} & 37.05\scriptsize{$\pm$0.00} \\
& \method &  60.91\scriptsize{$\pm$2.18}	& 57.71\scriptsize{$\pm$1.82}	& 65.23\scriptsize{$\pm$1.37}	& 84.23\scriptsize{$\pm$1.90}	& 21.77\scriptsize{$\pm$2.17}	& 49.24\scriptsize{$\pm$3.55}	& 51.09\scriptsize{$\pm$8.61}	& 69.85\scriptsize{$\pm$0.51} \\
\midrule
\multirow{2}{*}{\makecell{Few-Shot Labeling}} & \method-20\% & 49.04\scriptsize{$\pm$7.20} & 46.35\scriptsize{$\pm$4.28} & 57.48\scriptsize{$\pm$11.12} & 34.01\scriptsize{$\pm$6.45} & 9.38\scriptsize{$\pm$1.49} & 45.80\scriptsize{$\pm$3.41} & 45.39\scriptsize{$\pm$2.31}& 68.85\scriptsize{$\pm$2.62} \\
& \method-5 & 51.29\scriptsize{$\pm$4.41} & 46.67\scriptsize{$\pm$8.78}  & 57.92\scriptsize{$\pm$12.11} & 79.33\scriptsize{$\pm$5.38} & 21.56\scriptsize{$\pm$1.18} & 51.06\scriptsize{$\pm$0.86} & 40.2\scriptsize{$\pm$6.46}& 70.27\scriptsize{$\pm$4.51} \\
\bottomrule
\end{tabular}
\end{adjustbox}
\end{table*}

\begin{table*}[]
\caption{Per-dataset results of performance on zero-shot KG inductive reasoning.}
\small
\begin{adjustbox}{max width=\textwidth}
\begin{tabular}{l|cccccccccccccc}
\toprule
\multirow{2}{*}{\textbf{Datasets}} & \multicolumn{2}{c}{\textbf{Supervised SOTA}} & \multicolumn{2}{c}{\textbf{ULTRA(3g)}} & \multicolumn{2}{c}{\textbf{\method}} & \multicolumn{2}{c}{\textbf{\method(One)}} & \multicolumn{2}{c}{\textbf{\method(MPNet)}} & \multicolumn{2}{c}{\textbf{\method(Ontology)}} & \multicolumn{2}{c}{\textbf{\method(4g)}} \\
  & MRR & Hits@10 & MRR & Hits@10 & MRR & Hits@10 & MRR & Hits@10 & MRR & Hits@10 & MRR & Hits@10 & MRR & Hits@10\\
 \midrule
FB:v1 & 0.457 & 0.589 & 0.486 & 0.657 & 0.496 & 0.684 & 0.489 & 0.670 & 0.496 & 0.684 & 0.493 & 0.681 & 0.499 & 0.657\\
FB:v2 & 0.51 & 0.672 & 0.501 & 0.694 & 0.511 & 0.720 & 0.507 & 0.709 & 0.509 & 0.718 & 0.498 & 0.713 & 0.509 & 0.713\\
FB:v3 & 0.476 & 0.637 & 0.482 & 0.644 & 0.490 & 0.666 & 0.485 & 0.656 & 0.491 & 0.667 & 0.485 & 0.663 & 0.494 & 0.663\\
FB:v4 & 0.466 & 0.645 & 0.477 & 0.671 & 0.485 & 0.683 & 0.481 & 0.678 & 0.485 & 0.682 & 0.481 & 0.679 & 0.489 & 0.676\\
WN:v1 & 0.741 & 0.826 & 0.593 & 0.779 & 0.661 & 0.795 & 0.641 & 0.772 & 0.663 & 0.799 & 0.658 & 0.780 & 0.640 & 0.796\\
WN:v2 & 0.704 & 0.798 & 0.620 & 0.752 & 0.650 & 0.785 & 0.657 & 0.765 & 0.650 & 0.783 & 0.653 & 0.755 & 0.645 & 0.788\\
WN:v3 & 0.452 & 0.568 & 0.371 & 0.494 & 0.399 & 0.529 & 0.387 & 0.517 & 0.400 & 0.532 & 0.373 & 0.492 & 0.388 & 0.520\\
WN:v4 & 0.661 & 0.743 & 0.484 & 0.687 & 0.594 & 0.704 & 0.592 & 0.699 & 0.598 & 0.704 & 0.598 & 0.688 & 0.590 & 0.714\\
NL:v1 & 0.637 & 0.866 & 0.716 & 0.861 & 0.783 & 0.913 & 0.743 & 0.861 & 0.771 & 0.908 & 0.764 & 0.898 & 0.745 & 0.888\\
NL:v2 & 0.419 & 0.601 & 0.525 & 0.719 & 0.538 & 0.761 & 0.533 & 0.750 & 0.540 & 0.760 & 0.516 & 0.739 & 0.552 & 0.753\\
NL:v3 & 0.436 & 0.594 & 0.511 & 0.687 & 0.554 & 0.750 & 0.553 & 0.750 & 0.552 & 0.751 & 0.544 & 0.740 & 0.556 & 0.753\\
NL:v4 & 0.363 & 0.556 & 0.490 & 0.701 & 0.493 & 0.740 & 0.494 & 0.732 & 0.493 & 0.734 & 0.475 & 0.712 & 0.499 & 0.739\\
\midrule
FB:25 & 0.133 & 0.271 & 0.383 & 0.633 & 0.389 & 0.645 & 0.388 & 0.641 & 0.389 & 0.645 & 0.386 & 0.640 & 0.387 & 0.640\\
FB:50 & 0.117 & 0.218 & 0.330 & 0.536 & 0.341 & 0.548 & 0.335 & 0.537 & 0.341 & 0.549 & 0.336 & 0.541 & 0.340 & 0.543\\
FB:75 & 0.189 & 0.325 & 0.391 & 0.594 & 0.400 & 0.611 & 0.399 & 0.603 & 0.400 & 0.610 & 0.395 & 0.603 & 0.397 & 0.603\\
FB:100 & 0.223 & 0.371 & 0.438 & 0.631 & 0.437 & 0.642 & 0.438 & 0.636 & 0.438 & 0.640 & 0.431 & 0.637 & 0.439 & 0.642\\
WK:25 & 0.186 & 0.309 & 0.307 & 0.507 & 0.292 & 0.497 & 0.297 & 0.495 & 0.290 & 0.491 & 0.289 & 0.483 & 0.301 & 0.518\\
WK:50 & 0.068 & 0.135 & 0.158 & 0.296 & 0.160 & 0.299 & 0.159 & 0.295 & 0.159 & 0.299 & 0.146 & 0.293 & 0.173 & 0.318\\
WK:75 & 0.247 & 0.362 & 0.373 & 0.519 & 0.365 & 0.532 & 0.368 & 0.522 & 0.365 & 0.531 & 0.342 & 0.514 & 0.375 & 0.536\\
WK:100 & 0.107 & 0.169 & 0.178 & 0.289 & 0.186 & 0.302 & 0.176 & 0.283 & 0.186 & 0.302 & 0.142 & 0.289 & 0.188 & 0.309\\
NL:25 & 0.334 & 0.501 & 0.387 & 0.538 & 0.392 & 0.601 & 0.359 & 0.562 & 0.394 & 0.604 & 0.376 & 0.566 & 0.404 & 0.612\\
NL:50 & 0.281 & 0.453 & 0.398 & 0.549 & 0.394 & 0.565 & 0.375 & 0.540 & 0.394 & 0.567 & 0.381 & 0.557 & 0.406 & 0.589\\
NL:75 & 0.261 & 0.464 & 0.348 & 0.527 & 0.349 & 0.535 & 0.350 & 0.519 & 0.350 & 0.540 & 0.341 & 0.535 & 0.360 & 0.562\\
NL:100 & 0.309 & 0.506 & 0.442 & 0.631 & 0.475 & 0.695 & 0.468 & 0.692 & 0.473 & 0.693 & 0.464 & 0.678 & 0.476 & 0.687\\
\bottomrule
\end{tabular}
\end{adjustbox}
\end{table*}


\end{document}